%% file: main.tex
\documentclass{article} 
\usepackage{iclr2025_conference,times}

\input{macros/math_commands.tex}

\usepackage{hyperref}
\hypersetup{
    colorlinks=false,
    linkcolor=red,
    filecolor=magenta,      
    urlcolor=cyan,
}
\usepackage{url}
\usepackage{amsmath}
\usepackage{enumitem}
\usepackage{tikz}
\usetikzlibrary{arrows.meta, positioning, shapes, calc, matrix, backgrounds, fit}
\usepackage{dsfont}
\usepackage{subfigure}
\usepackage{siunitx}
\usepackage[acronym]{glossaries}
\usetikzlibrary{positioning}
\usepackage[makeroom]{cancel}
\usepackage{caption}

\input{macros/notations}

\title{DDPS: Discrete Diffusion Posterior Sampling \\for Paths in Layered Graphs}

\author{
Hao~Luan$^1$,   
See-Kiong~Ng$^{1,2}$,  and 
Chun~Kai~Ling$^1$\\
$^1$School of Computing, National University of Singapore\\
$^2$Institute of Data Science, National University of Singapore\\
\texttt{haoluan@comp.nus.edu.sg, \{seekiong, chunkail\}@nus.edu.sg} \\
}

\input{macros/gloss}

\iclrfinalcopy 
\begin{document}

\maketitle

\begin{abstract}
Diffusion models form an important class of generative models today, accounting for much of the state of the art in cutting edge AI research. While numerous extensions beyond image and video generation exist, few of such approaches address the issue of \textit{explicit constraints} in the samples generated. In this paper, we study the problem of generating paths in a layered graph (a variant of a directed acyclic graph) using discrete diffusion models, while guaranteeing that our generated samples are indeed paths. Our approach utilizes a simple yet effective representation for paths which we call the padded adjacency-list matrix (PALM). In addition, we show how to effectively perform classifier guidance, which helps steer the sampled paths to specific preferred edges without any retraining of the diffusion model. Our preliminary results show that empirically, our method outperforms alternatives which do not explicitly account for path constraints.
\end{abstract}

\input{sections/1_intro}

\input{sections/2_1_formulation}
\input{sections/3_palm}

\input{sections/4_guidance}
\input{sections/5_experiment}

\newpage
\bibliography{ref}
\bibliographystyle{iclr2025_conference}

\newpage
\input{sections/appendices}

\end{document}

%% file: macros/math_commands.tex
\usepackage{amsmath,amsfonts,bm}

\def\figref#1{figure~\ref{#1}}
\def\Figref#1{Figure~\ref{#1}}

\def\secref#1{section~\ref{#1}}

\def\eqref#1{Eq.~(\ref{#1})}

\def\1{\bm{1}}

\DeclareMathAlphabet{\mathsfit}{\encodingdefault}{\sfdefault}{m}{sl}
\SetMathAlphabet{\mathsfit}{bold}{\encodingdefault}{\sfdefault}{bx}{n}

\newcommand{\R}{\mathbb{R}}



%% file: macros/notations.tex
\usepackage{amsthm,amsmath,amssymb,amscd,mathrsfs}
\usepackage{txfonts,amsfonts,bbm}
 
\usepackage{color}
\usepackage{algorithm}
\usepackage[noend]{algpseudocode}
\usepackage{hyperref}
\hypersetup{
    colorlinks=true,
    linkcolor=red,
    filecolor=magenta,      
    urlcolor=blue,
}
\usepackage{xspace}

\theoremstyle{definition}
\newtheorem{defn}{Definition}
\theoremstyle{remark}
\newtheorem{rmk}{Remark}

\newcommand{\etc}{\textit{etc.}\xspace}
\newcommand{\eg}{{\it e.g.}}
\newcommand{\ie}{{\it i.e.}}

\newcommand{\perse}{\textit{per~se}\xspace}

\newcommand{\dfnref}[1]{{\bf Definition~\ref{#1}}}

\newcommand{\algoref}[1]{\textbf{Algorithm~\ref{#1}}}

\newcommand{\ptar}[1]{p_{\mathrm{tar}}\left( {#1} \right)}
\newcommand{\pgen}[1]{p_{\mathrm{gen}}\left( {#1} \right)}

\newcommand{\condpdf}[3]{#1\left( {#2}\mid{#3}\right)}

\newcommand{\myeexpect}[2]{\mathop{\mathbb{E}}_{#1} \left[ #2 \right]}

\newcommand{\kldiv}[2]{\mathop{\mathbb{D}_{\text{KL}}} \left[ #1 \,\|\, #2\right]}

\newcommand{\ggrad}[2]{\nabla_{#1}{#2}}

\usepackage{ulem} 

\newcommand{\I}{\boldsymbol{I}}

\PassOptionsToPackage{normalem}{ulem}

\usepackage{hyperref} 
\usepackage{enumitem}
\usepackage{float}
\usepackage{setspace}
\usepackage{graphicx}

\floatstyle{ruled}
\newfloat{algorithm}{tbp}{loa}
\providecommand{\algorithmname}{Algorithm}
\floatname{algorithm}{\protect\algorithmname}

%% file: macros/gloss.tex
\setacronymstyle{long-short}

\makenoidxglossaries

\newacronym{dm}{DM}{diffusion model}
\newacronym{ddpm}{DDPM}{Denoising Diffusion Probabilistic Models}
\newacronym{sgm}{SGM}{Score-based Generative Model}

\newacronym{sde}{SDE}{stochastic differential equation}
\newacronym{ode}{ODE}{ordinary differential equation}
\newacronym{vae}{VAE}{variational Autoencoder}

\newacronym{palm}{PALM}{padded adjacency--list matrix}
\newacronym{fid}{FID}{Fr\'echet Inception distance}
\newacronym{flgd}{FLGD}{Fr\'echet layered graph distance}
\newacronym{isl}{IS-L}{Layer Imitation Score}

\newcommand{\tvd}[2]{\mathop{d_{\mathrm{TV}}}\left[ {#1},\, {#2}\right]}

\newcommand{\sfdord}[3]{\mathrm{SFD}^{#1} \left[ {#2},\, {#3}\right]}
\newcommand{\isl}{IS-L}
\newcommand{\islsf}{IS-L-SF\xspace}
\newcommand{\isllone}{IS-L-$L^1$}
\newcommand{\islkl}{IS-L-KL}
\newcommand{\isltv}{IS-L-TV}

\newcommand{\method}{\textsc{DDPS}\xspace}

\newcommand{\dddpm}{\textsc{D3PM}\xspace}
\newcommand{\swingnn}{\textsc{SwinGNN}\xspace}
\newcommand{\edpgnn}{\textsc{EDP-GNN}\xspace}

%% file: sections/1_intro.tex
\section{Introduction}%
\label{sec:introduction}
Diffusion models have emerged as one of the most popular methods of generative AI particularly with hyper-realistic image and video generation, often outperforming older methods like generative adversarial networks. 
The recent years have seen much interest in replicating this success in other domains. These domains include protein design \citep{frey2024protein}, molecular conformations \citep{xu2022geodiff}, text generation \citep{li2022diffusion}, robotics \citep{chi2023diffusion,wang2024equivariant,feng2024ltldog,feng2025diffusion}, \etc  
Unlike image and video generation, These recent applications often require the restriction that the generated samples belong to some \textit{discrete domain}. 
On top of that, one often requires samples to obey some form of predefined \textit{structural constraint}, either for reasons related to safety or simply because violating those constraints would make little physical sense. For instance, certain robot poses or physical configurations cannot be achieved in the real world. Most of the existing work adopt a ``data-centric'' approach towards ``softly'' enforcing such constraints \citep{chi2023diffusion}, relying on the assumption that such unsafe or non-physical samples do not appear in the training set and that these structural constraints would be implicitly learned. 

In this paper, we study how to \textit{explicitly} enforce these structural constraints. We focus on trying to generate paths in a \textit{layered graph}, which are directed acyclic graphs organized in layers. Layered graphs occur regularly in domains involving planning. 
For this reason, we seek to be able to sample from such paths (learned from some dataset) via a diffusion model. This has applications ranging from security to logistics\citep{cerny2024layered,vcerny2024contested,zhang2017optimal}. However, this requires that the generated samples from the diffusion model obey the constraint that they are indeed valid paths in the layered graph, and not just an arbitrary subset of edges/vertices.

Our contributions are as follows. First, we formulate the path learning problem in a layered graph via discrete diffusion models. We propose a simple representation of paths called \gls{palm}, which guarantees that generated output will always be paths in the underlying layered graph, unlike other diffusion based approaches. Second, we show how to perform training and inference efficiently under \gls{palm}. Third, we show how one can perform \textit{guidance} under our proposed model, a variant of existing classifier guidance methods that allow us to favor paths which contain certain distinguished edges during inference without retraining. Lastly, we show empirically that our proposed method is superior in generating paths compared to naive alternatives. We also examine the tradeoff between strength of guidance and adherence to the learned distribution.

%% file: sections/2_1_formulation.tex
\section{Preliminaries and Related Work}
    In this paper, we work with paths in \textit{layered graphs}. Loosely speaking, a layered graph is a directed acyclic graph with the additional structure that the vertices can be ordered by layers \citep{cerny2024layered}; see \figref{fig:lg_example} for an example.
    Our goal is to learn and generate paths in a layered graph with a diffusion model. Furthermore, we would like the generated paths to possess some user-specified properties, \eg, passing through some preferred edges. As we will see later, this will be achieved by employing discrete diffusion models and classifier guidance based posterior sampling. 

    \subsection{Preliminaries}
    \begin{defn}[Layered Graph, adapted from~\citep{cerny2024layered}]%
    \label{dfn:lg}
        Let $G = (\mathcal V, \mathcal E)$ be a \textit{directed} graph defined over a finite vertex set $\mathcal V$ and edge set $\mathcal E$. 
        $G$ is a layered graph if all of the following conditions are true: 
        (1) Set $\mathcal V$ can be partitioned into $L>1$ non-empty sets $\mathcal V^1,\ldots,\mathcal V^{L}$, which are called \textit{layers}; 
        (2) each edge $e\in \mathcal E$ is in $\mathcal V^{l} \times \mathcal V^{l+1}$ for some $l\in [1, L-1]$; 
        (3) the first layer is a singleton: $|\mathcal V^1|=1$; 
        (4) every vertex $v\in \mathcal V^l$ for $l<L$ with zero out-degree has zero in-degree. 
    \end{defn}
    
    \begin{defn}[Path]%
    \label{dfn:lg_path}
        A path in a layered graph is defined as an ordered sequence 
        $(v^{1}, \ldots, v^{L}) \in \mathcal V^1 \times \cdots \times \mathcal V^L$ 
        of length $L$, 
        such that $(v^l, v^{l+1}) \in \mathcal E$ for $l=1,\ldots,L-1$. 
    \end{defn}

    \textbf{Examples of layered graphs.} In the work of \citeauthor{cerny2024layered}, each layer represented the possible location of an agent (typically a security patrol) at a given time; the existence of an edge between adjacent layers meant that moving between two physical locations is possible at a particular timestep. Hence, each valid path represents a different duration route $L$. Another example is sequence prediction. For instance, a fully connected layered graph with four vertices per layer can be used to represent all possible combinations of length $L$ DNA sequences. Here, edges can be used to forbid certain adjacent nucleotide bases (\eg, ``A'' cannot be followed by ``G'' in the 4th and 5th location).
    
    \begin{table}[t]
        \centering
        \begin{minipage}{0.45\textwidth}
                \input{figures/tikz_lg}
                
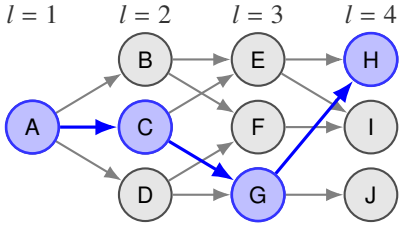
\captionof{figure}{
                    Example of a layered graph with $L=4$ layers. 
                    The first layer $\mathcal{V}^1$ is a singleton ($|\mathcal{V}^1|=1$), while subsequent layers have $|\mathcal{V}^l|=3$ for $l=2,3,4$. 
                    Edges only exist between adjacent layers ($l \to l+1$). 
                    A specific path (A, C, G, H) is highlighted in blue.
                }
                \label{fig:lg_example}
        \end{minipage}
        \hfill
        \begin{minipage}{0.52\textwidth}
            \input{figures/tikz_palm}
            \captionof{figure}{
                The \gls{palm} representation (transposed) for the path highlighted in \figref{fig:lg_example}. 
                Columns correspond to vertices: blue indicates on-path vertices (A, C, G, H) and their specific one-hot vectors; gray indicates off-path vertices. 
                Each gray column vector is an arbitrary one-hot selection if the vertex has outgoing edges. 
                Faded entries are padding zeros. 
            }
            \label{fig:palm_example}
        \end{minipage}
    \vspace{-10pt}
    \end{table}

    \paragraph{Discrete diffusion models.}
    A discrete diffusion model involves with a forward Markov process gradually corrupting the initial discrete data representation $x_0$ with noise for a finite number of time steps $t=1, \ldots,T$,  and a learned reverse Markov process gradually reconstructing $x_0$ from $x_T$. 
    A single step in the forward chain can be formalized as 
    \begin{align}
        \condpdf{q}{x_t}{x_{t-1}} = \operatorname{Cat}(Q_t x_{t-1})
    \end{align}
    wherein 
    $\operatorname{Cat}(p)$ is a $K$-dimensional categorical distribution parameterized by the probabilities in $p \in \R^K$, 
    $x_t\in \R^K$ is a one-hot vector, 
    and $Q_t \in \R^{K \times K}$ is a transition matrix at timestep $t$ specifying transition probabilities among the $K$ categories.
    The diffusion model learns a backward process 
    \begin{align}
    \label{eq:d3pm_bkwd}
        \condpdf{p_\theta}{x_{t-1}}{x_t} = \sum_{x_0} \condpdf{q}{x_{t-1}}{x_t, x_0} \condpdf{p_\theta}{x_0}{x_t}
    .\end{align}
    At inference time, clean samples can be generated by running the learned backward process \eqref{eq:d3pm_bkwd} for multiple timesteps $t=T,\ldots,1$, with $x_T \sim p_T(x)$ sampled from some prior distribution $p_T$.

\subsection{Related Work}

Diffusion models \citep{sohl2015deep,ho2020denoising,song2019generative,song2021scorebased} first gained success in tasks in continuous domains like image generation~\citep{dhariwal2021diffusion,rombach2022high}. 
Influenced by such success, more approaches have been proposed for generating discrete data, \eg, text \citep{li2022diffusion}, graphs \citep{niu2020permutation,vignac2023digress,yan2024swingnn}, biological sequences \citep{yim2023se3,avdeyev2023dirichlet}, \etc  
Diffusion models operating on discrete state-space \citep{austin2021structured, hoogeboom2021argmax} are of our interest. 
\citet{campbell2022continuous,campbell2024generative} also extended discrete diffusion models from discrete-time setting to continuous time through the lens of Continuous-time Markov chain. 
There are also efforts in diffusion models working on the continuous space of probabilistic simplex \citep{stark2024dirichlet}.

A favorable trait of diffusion models is their amenability to guidance \citep{dhariwal2021diffusion,ho2021classifierfree}, which allows users to generate samples with desirable properties and has been widely exploited in tasks on continuous domains. 
For discrete diffusion, the technique of guidance is less mature. 
One of the pioneer attempts might be \citet{vignac2023digress} using guidance in graph generation. 
More discrete guidance techniques were recently proposed for biological sequence \citep{nisonoff2024unlocking}, text generation\citep{schiff2024simple} and for discrete latent diffusion models \citep{han2024guided, murata2024g2d2}. 
While most existing works approach guidance from the perspective of either classifier guidance \citep{dhariwal2021diffusion} or classifier-free guidance \citep{ho2021classifierfree}, recent emerging research explores a reinforcement learning approach \citep{rector2024steering,li2024derivative,wang2024fine,uehara2025inference}. 

To the best of our knowledge, there has not been much work on diffusion models that can guarantee that \textit{paths} are generated with \textit{certainty} without significant post processing. For instance, \citet{niu2020permutation} and \citet{yan2024swingnn} learn continuous representations of graph adjacency matrices and perform discretization as a form of post processing to obtain a graph --- this graph is not guaranteed to be a path, even if the training data is so. More closely related to us is the work of \citep{shigraph}, which approach path planning problems by generating sequences of vertices to traverse. However, there is no guarantee that vertices at adjacent timesteps are adjacent to each other, and the method relies on a non-trivial beam-search post-processing algorithm in order to guarantee path validity.

%% file: figures/tikz_lg.tex
\begin{tikzpicture}[
        >=Latex, 
        node_style/.style={
            circle, 
            draw=black!70, 
            fill=black!10, 
            thick, 
            minimum size=7mm, 
            font=\small\sffamily 
        },
        path_node/.style={
            node_style, 
            draw=blue!80, 
            fill=blue!25, 
            line width=1pt 
        },
        edge_style/.style={
            ->, 
            thick, 
            draw=black!50 
        },
        path_edge/.style={
            edge_style, 
            draw=blue, 
            line width=1.2pt 
        },
        layer_label/.style={
            font=\bfseries, 
            text=black!80
        }
      ]

      \coordinate (L1) at (0,0);
      \coordinate (L2) at (1.5cm, 0); 
      \coordinate (L3) at (3.0cm, 0); 
      \coordinate (L4) at (4.5cm, 0);

      \coordinate (LabelPosY) at (0, 1.5cm); 

      
      \node[node_style] (v11) at (L1) {A}; 
      
      \node[node_style] (v21) at ([yshift=0.9cm]L2) {B};  
      \node[node_style] (v22) at (L2) {C}; 
      \node[node_style] (v23) at ([yshift=-0.9cm]L2) {D}; 

      \node[node_style] (v31) at ([yshift=0.9cm]L3) {E};
      \node[node_style] (v32) at (L3) {F};
      \node[node_style] (v33) at ([yshift=-0.9cm]L3) {G};

      \node[node_style] (v41) at ([yshift=0.9cm]L4) {H};
      \node[node_style] (v42) at (L4) {I};
      \node[node_style] (v43) at ([yshift=-0.9cm]L4) {J};

      \begin{scope}[style=edge_style]
          \path (v11) edge (v21); 
          \path (v11) edge (v23); 
          
          \path (v21) edge (v31); 
          \path (v21) edge (v32); 
          \path (v22) edge (v31); 
          \path (v23) edge (v32); 
          \path (v23) edge (v33); 
          
          \path (v31) edge (v41); 
          \path (v31) edge (v42); 
          \path (v32) edge (v42); 
          \path (v33) edge (v43); 
      \end{scope}

      \node[path_node] (v11) at (L1) {A}; 
      \node[path_node] (v22) at (L2) {C};
      \node[path_node] (v33) at ([yshift=-0.9cm]L3) {G}; 
      \node[path_node] (v41) at ([yshift=0.9cm]L4) {H}; 

      \begin{scope}[style=path_edge]
          \path (v11) edge (v22); 
          \path (v22) edge (v33); 
          \path (v33) edge (v41); 
      \end{scope}

      \node[layer_label] at (L1 |- LabelPosY) {$l=1$}; 
      \node[layer_label] at (L2 |- LabelPosY) {$l=2$};
      \node[layer_label] at (L3 |- LabelPosY) {$l=3$};
      \node[layer_label] at (L4 |- LabelPosY) {$l=4$};

\end{tikzpicture}

%% file: figures/tikz_palm.tex
\begin{tikzpicture}
    \colorlet{onPathColor}{blue!20}    
    \colorlet{offPathColor}{gray!30}   
    \colorlet{paddingColor}{gray!60}  
    \colorlet{textColor}{black}
    \colorlet{gridColor}{gray!20}
    \colorlet{bracketColor}{black!70} 

    \tikzset{
        matrix cell/.style={ 
            rectangle,
            minimum size=5mm,
            anchor=center,
            draw=gridColor,
            font=\ttfamily,
        },
        row label/.style={ 
            minimum width=15mm,
            minimum height=8mm,
            anchor=east,
            font=\footnotesize,
            inner xsep=2mm,
        },
        col label/.style={ 
            minimum width=8mm,
            anchor=south,
            font=\bfseries,
            inner ysep=2mm,
        },
        padding text/.style={ 
            text=paddingColor,
        },
        active one/.style={ 
            font=\bfseries\ttfamily,
            text=textColor,
        },
        active zero/.style={ 
             text=textColor,
        },
    }

    \matrix (palm_matrix) [
        matrix of nodes,
        nodes={matrix cell},
        row sep=-\pgflinewidth,
        column sep=-\pgflinewidth,
        nodes in empty cells,
        row 1 column 1/.style={nodes={active zero}}, row 2 column 1/.style={nodes={active one}}, row 3 column 1/.style={nodes={active zero}},
        row 1 column 2/.style={nodes={active one}}, row 2 column 2/.style={nodes={active zero}}, row 3 column 2/.style={nodes={padding text}},
        row 1 column 3/.style={nodes={active zero}}, row 2 column 3/.style={nodes={active one}}, row 3 column 3/.style={nodes={padding text}},
        row 1 column 4/.style={nodes={active zero}}, row 2 column 4/.style={nodes={active one}}, row 3 column 4/.style={nodes={padding text}},
        row 1 column 5/.style={nodes={active zero}}, row 2 column 5/.style={nodes={active one}}, row 3 column 5/.style={nodes={padding text}},
        row 1 column 6/.style={nodes={active one}}, row 2 column 6/.style={nodes={padding text}}, row 3 column 6/.style={nodes={padding text}},
        row 1 column 7/.style={nodes={active one}}, row 2 column 7/.style={nodes={active zero}}, row 3 column 7/.style={nodes={padding text}},
        row 1 column 8/.style={nodes={padding text}}, row 2 column 8/.style={nodes={padding text}}, row 3 column 8/.style={nodes={padding text}},
        row 1 column 9/.style={nodes={padding text}}, row 2 column 9/.style={nodes={padding text}}, row 3 column 9/.style={nodes={padding text}},
        row 1 column 10/.style={nodes={padding text}}, row 2 column 10/.style={nodes={padding text}}, row 3 column 10/.style={nodes={padding text}},
    ]
    { 
      0 & 1 & 0 & 0 & 0 & 1 & 1 & 0 & 0 & 0 \\
      1 & 0 & 1 & 1 & 1 & 0 & 0 & 0 & 0 & 0 \\
      0 & 0 & 0 & 0 & 0 & 0 & 0 & 0 & 0 & 0 \\ 
    };

    \foreach \col/\label in {1/A, 2/B, 3/C, 4/D, 5/E, 6/F, 7/G, 8/H, 9/I, 10/J} {
        \node[col label] at (palm_matrix-1-\col.north) {\label};
    }

    \foreach \row/\label [count=\i] in {1/Edge 1, 2/Edge 2, 3/Edge 3} {
        \node[row label] at (palm_matrix-\row-1.west) {\label};
    }

    \begin{scope}[on background layer]
        \node[fit=(palm_matrix-1-1) (palm_matrix-3-1), fill=onPathColor, inner sep=0pt] {}; 
        \node[fit=(palm_matrix-1-3) (palm_matrix-3-3), fill=onPathColor, inner sep=0pt] {}; 
        \node[fit=(palm_matrix-1-7) (palm_matrix-3-7), fill=onPathColor, inner sep=0pt] {}; 
        \node[fit=(palm_matrix-1-8) (palm_matrix-3-8), fill=onPathColor, inner sep=0pt] {}; 
        \node[fit=(palm_matrix-1-2) (palm_matrix-3-2), fill=offPathColor, inner sep=0pt] {}; 
        \node[fit=(palm_matrix-1-4) (palm_matrix-3-4), fill=offPathColor, inner sep=0pt] {}; 
        \node[fit=(palm_matrix-1-5) (palm_matrix-3-5), fill=offPathColor, inner sep=0pt] {}; 
        \node[fit=(palm_matrix-1-6) (palm_matrix-3-6), fill=offPathColor, inner sep=0pt] {}; 
        \node[fit=(palm_matrix-1-9) (palm_matrix-3-9), fill=offPathColor, inner sep=0pt] {}; 
        \node[fit=(palm_matrix-1-10) (palm_matrix-3-10), fill=offPathColor, inner sep=0pt] {}; 
    \end{scope}

    \def\bracketOffset{1pt} 
    \def\bracketExtend{1pt} 
    \def\bracketTick{3pt}   

    \coordinate (TL) at ($(palm_matrix.north west)+(-\bracketOffset, \bracketExtend)$);
    \coordinate (BL) at ($(palm_matrix.south west)+(-\bracketOffset,-\bracketExtend)$);
    \coordinate (TR) at ($(palm_matrix.north east)+(\bracketOffset, \bracketExtend)$);
    \coordinate (BR) at ($(palm_matrix.south east)+(\bracketOffset,-\bracketExtend)$);

    \draw[line width=1.5pt, color=bracketColor]
        (TL) --++ (\bracketTick,0) 
        (TL) -- (BL) 
        (BL) --++ (\bracketTick,0); 

    \draw[line width=1.5pt, color=bracketColor]
        (TR) --++ (-\bracketTick,0) 
        (TR) -- (BR) 
        (BR) --++ (-\bracketTick,0); 

    \node[anchor=south west, font=\large\bfseries, color=bracketColor] at ($(TR)+(\bracketTick-2pt, -2pt)$) {$\mathsf{T}$};

\end{tikzpicture}

%% file: sections/3_palm.tex
\section{Structured Discrete Representation of Paths}%
\label{sec:palm}
    Choosing an appropriate data representation of paths is important for learning and sampling with diffusion models. 
    A direct and naive approach is to treat each path as a subgraph of the original layered graph and learn the distribution of these sub-graphs with a diffusion model. 
    This approach may be implemented using standard continuous diffusion models to learn a \textit{continuous relaxation} of the adjacency matrix of a graph \citep{niu2020permutation,yan2024swingnn}, the output of which may then be truncated or rounded off to yield a sampled subgraph.
    However, as we will show in \secref{sub:discrete_quality}, such an approach results in the degradation of sample quality--- many subgraphs do not correspond to paths in the layered graph.
    The key drawback in this naive approach is that it is oblivious to the underlying \textit{structural constraints} distinguishing a path from a general (sub-)graph and instead relies on this ``pattern'' in the data to be learned by the diffusion model itself. 
    Our \gls{palm} representation seeks to overcome this issue by explicitly perform learning on a representation that maps to paths. 
    \begin{defn}[Padded Adjacency--List Matrix (PALM)]%
    \label{dfn:palm}
        Given a layered graph $G=(\mathcal V, \mathcal E)$ with $L$ layers, let $V=|\mathcal V|$ and $\mathcal V=\{1, \ldots, V\}$.  
        $D_v=\operatorname{deg}(v)$, where $\operatorname{deg}(\cdot)$ refers to the out-degree of a vertex. 
        The edge set $\mathcal E$ can also be partitioned into exactly $V$ sets $\{\mathcal E_1,\ldots,\mathcal E_V\}$, where each edge $e\in \mathcal E_v$ originates at vertex $v$. 
        For a path $(v^1, \cdots, v^L)$, its \gls{palm} representation is a collection of vectors $\{ x^1, \ldots, x^V\}$ where $x^v \in \R^{D_v}$ is a one-hot vector indicating an edge $e\in \mathcal E_v$ for all $v\in \mathcal V$. 
    \end{defn}

    At its core, a \gls{palm} is a stack of one-hot vectors recording a single outgoing edge of each vertex, as demonstrated in \figref{fig:palm_example}. 
    It is easy to see that the \gls{palm}-to-path mapping is many-to-one and onto, which means (\expandafter{\romannumeral 1}) one \gls{palm} instance represents \textit{exactly} one path, and (\expandafter{\romannumeral 2}) all paths can be represented by at least one \gls{palm}.  
    The path-to-\gls{palm} conversion is trivial by \dfnref{dfn:palm}; 
    the \gls{palm}-to-path conversion is intuitive: 
    starting by $v^l=v^1$ (which must be the first vertex appearing in any path per \dfnref{dfn:lg}), 
    follow the edge $e=(v^l, v^{l+1})$ represented by $x^l$ in the \gls{palm} and transit to $v^{l+1}$, and then repeat this process until reaching $v^L$.

    \paragraph{Diffusion learning and inference with PALM.}
        We employ the \dddpm framework of \citep{austin2021structured} to learn a \gls{palm} representation. 
        We conduct training and inference with \gls{palm} as the representation of paths. 
        Concretely, a neural network is trained to take \gls{palm} as input and predict the logits of a distribution: $\condpdf{\tilde p_\theta}{\tilde x_0}{x_t}$. 
        We adopt the same parametrization as \citet{austin2021structured}:  
        \begin{align}
            \condpdf{p_\theta}{x_{t-1}}{x_t} \propto 
            \sum_{\tilde x_0} \condpdf{q}{x_{t-1}, x_t}{\tilde x_0} 
            \condpdf{\tilde p_\theta}{\tilde x_0}{x_t}
        .\end{align}
        To learn path distributions, we make the following changes from \dddpm. First, since our dataset only stores paths but not \gls{palm}, we convert paths to \gls{palm} on the fly during training. 
        Since the \gls{palm}-to-path mapping is many-to-one, we assign one hot vectors uniformly in rows (corresponding to vertices in the layered graph) that \textit{do not} belong to the path.
        We adopt the uniform transition matrix \citep{austin2021structured,hoogeboom2021argmax} in the forward diffusion process,  
        but the transition matrix at each timestep $t$ for each node $v$ is constructed differently according to their out-degrees, catering the varied length of vectors in \gls{palm} (see details in Appendix~\ref{apdx:expt}). 
        We use the common cosine noise scheduling \citep{nichol2021improved} for the forward diffusion process. 
        For training, we follow the combined loss proposed by \citet{austin2021structured}: 
        \begin{align}
            L_\gamma = \gamma L_{\text{vb}} + \myeexpect{q(x_0)}{\myeexpect{\condpdf{q}{x_t}{x_0}}{-\log \condpdf{p_\theta}{\tilde x_0}{x_t}}}
        ,\end{align}
        where $L_{\text{vb}}$ is the variation bound loss commonly adopted in diffusion models~\citep{sohl2015deep,ho2020denoising,austin2021structured} and $\gamma$ is a weighting parameter.
        Since we are learning distributions over paths rather than \gls{palm}, we only include losses incurred at vertices \textit{on the path}.

        We perform inference (sampling of paths) in \gls{palm} representation the same way as \dddpm, with one extra step at the end that converts the sampled paths $x_0$ in \gls{palm} representation back to the common sequence form defined in \dfnref{dfn:lg_path}. Our implementation of \dddpm is adapted from \citep{d3pm_pytorch} and  \citep{niu2020permutation,yan2024swingnn}.

%% file: sections/4_guidance.tex
\section{Discrete Diffusion Guidance via Posterior Sampling}%
\label{sec:ddps}

    Guidance \citep{dhariwal2021diffusion, ho2021classifierfree} is a useful technique in diffusion-based conditional generation for steering the sampling process to data modes with the desired properties, \eg, a certain class of pictures, or in our case, a collection of paths that can achieve high rewards under a reward function. 
    In this work, we focus on the method of classifier guidance \citep{dhariwal2021diffusion,nichol2022glide} for compatibility to different external conditions (\eg, different reward function instances) \textit{without} the need for retraining the diffusion model. 
        
    Concretely, our objective of guided sampling is to generate paths that contain some or all of a set of prespecified ``preferred'' edges. 
    This defines an implicit scalar reward function $r$ over all possible paths, \ie, the reward associated with a path is equal to the number of preferred edges it contains.  

    \subsection{Guidance in Discrete Diffusion} 
    There has not been an established principle for guidance in discrete diffusion. 
    Unlike its counterpart in continuous domain where guidance under a given condition $y$ can be interpreted as a combination of the Stein score $\ggrad{x}{\log p(x)}$ and the likelihood score $\ggrad{x}{\condpdf{p}{y}{x}}$. 
    In discrete domains the (Stein) score is undefined, rendering this approach unfounded. 
    The discrete generalization of the score \citep{meng2022concrete,lou2024discrete}, on the other hand, has not yet provided a principled way for guidance either. 
    As such, we take a step back and return to the (logarithmic) Bayes' theorem \perse to derive the posterior in discrete diffusion as recent efforts \citep{nisonoff2024unlocking} have attempted:
    \begin{align}
        \underbrace{\log \condpdf{p}{x_{t-1}}{x_t, y}}_{\text{posterior}}
        =&  \underbrace{\log \left( 
                \condpdf{p}{y}{x_t, x_{t-1}}/
                \condpdf{p}{y}{x_t}
            \right)}_{\text{likelihood ratio}} 
            + \underbrace{
                \log \condpdf{p}{x_{t-1}}{x_t}
                }_{\text{prior}}\label{eq:bayes}
    ,\end{align}
    where the prior $\condpdf{p}{x_{t-1}}{x_t}$ is approximated by a trained denoising diffusion model $\condpdf{p_\theta}{x_{t-1}}{x_t}$. 
    Different yet akin to guidance in the continuous domain, the key to the posterior boils down to a \textit{reasonable} and \textit{efficient} approximation or direct computation of the log likelihood ratio 
    $\log\frac{
       \condpdf{p}{y}{x_t, x_{t-1}}
     }{
     \condpdf{p}{y}{x_t}
     }$.

    \subsection{Guidance Signal with PALM}
    \label{sub:guidance_signal}

        Given a scalar reward function $r$  
        we formulate condition $y$ in terms of reward optimality to influence edge selection probabilities. While directly increasing the sampling probabilities of preferred edges appears intuitive, this approach is inadequate since edge selection depends on the choices made in preceding layers.
        Consider a preferred edge $e = (v^l, v^{m})$ in a path $p$. When its preceding edges $e^\prime = (v^k, v^n)$, where $k < l$, have negligible selection probabilities, path $p$ becomes effectively inaccessible regardless of the selection probability of edge $e$. 
       
        Thus, we opt for utilizing the expectation of total rewards $\overline R$ given a distribution of \gls{palm}, involving all the probabilities of selecting each edge in the graph.  
        These probabilities are all encoded in the categorical distributions' logits $z$ yielded by the diffusion model $\condpdf{\tilde p_\theta}{\tilde x_0}{x_t}$. 
        Further, we propose to leverage the gradient of the expected total rewards with respect to the logits $z$, to approximate the log likelihood ratio as guidance signal: 
        \begin{align}
        \label{eq:likelihood_approx}
            \log \left( 
                \condpdf{p}{y}{x_t, x_{t-1}} /
                \condpdf{p}{y}{x_t}
            \right)
            \approx
            \ggrad{z}{\overline{R}(z)}
        .\end{align}
        Approximation (\ref{eq:likelihood_approx}) results in the reward-guided sampling process
        \begin{equation}
        \label{eq:guided_sampling}
            \log \condpdf{p_\theta}{x_{t-1}}{x_t, y}
            =  \lambda \ggrad{z}{\overline{R}(z)} 
                + \log \condpdf{p_\theta}{x_{t-1}}{x_t}, 
                \quad \text{where} \quad z = \condpdf{\tilde p_\theta}{\tilde x_0}{x_t}
        .\end{equation}
        
         \algoref{alg:ddps} describes this guided sampling procedure. 

        \begin{algorithm}
        \caption{Total Expected Reward Calculation}
        \label{alg:expect_reward}
        \begin{algorithmic}[1]
            \State \textbf{Inputs:}
                \gls{palm} Logits $z$, 
                Layered graph $G=(\mathcal V, \mathcal E)$, 
                Reward assignment \gls{palm} $[u]$
            \State $[p_v] \leftarrow \texttt{get\_transit\_prob\_dp}(G, z)$
                \Comment{Get transition prob. to each node $v$}
            \For{$v \in \mathcal V$}
                \State 
                    $r_v \leftarrow p_v \cdot \mathbf{1}^\top u_v $ 
            \EndFor
            \State $\overline{R} \leftarrow \sum_{v \in V} r_v$ 
            \State \textbf{Return:} $\overline{R}$
        \end{algorithmic}
        \end{algorithm} 
        
        \begin{algorithm}
        \caption{\method Inference}
        \label{alg:ddps}
        \begin{algorithmic}[1]
            \State \textbf{Inputs:}
                Diffusion model $\hat p_{\theta}$
                Reward model $\overline{R}$, 
                Forward noise $Q_1,\cdots, Q_T$,
                Guidance scale $\lambda$
            \State $x_T \sim \operatorname{Cat}(\mathbf{1})$ 
                \Comment{Sample from uniform categorical distribution}
            \For{$t=T, T-1, \ldots, 1$}
                \State $z \leftarrow \tilde p_\theta(\tilde x_0 | x_t)$
                    \Comment{Predict $x_0$ logits with diffusion model}
                \State $\bar r \leftarrow \overline{R}(z)$
                    \Comment{Get expected rewards via \algoref{alg:expect_reward}}
                \State $g_t \leftarrow \ggrad{z}{\bar{r}}$
                \State $\tilde x_{t-1} \leftarrow (\prod_{\tau=1}^t Q_\tau) z$
                    \Comment{Forward diffusion process}
                \State $\tilde x_{t-1}^\prime \leftarrow \lambda g_t + \tilde x_{t-1}$
                    \Comment{Perform guidance to get posterior}
                \State 
                    $x_{t-1} \sim \operatorname{Cat}(\tilde x_{t-1}^\prime)$ 
                    \Comment{Sample from the guided posterior}
            \EndFor
            \State \textbf{Return:} $x_0$
        \end{algorithmic}
        \end{algorithm}
        \vspace*{5pt}

%% file: sections/5_experiment.tex
\section{Experiments}%
\label{sec:experiments}

In this section, we address the following research questions: 
\vspace*{-5pt}
\begin{enumerate}[start=1,label={\bfseries RQ\arabic*:},leftmargin=*]
    \item In terms of quality of generated samples, how does our proposed \gls{palm} representation compared to alternatives like adjacency matrix learning \citep{niu2020permutation,yan2024swingnn}?
    \item Is the proposed \method method effective for reward improvement? 
    \item 
        When utilizing guidance, what is the tradeoff between reward improvement and adherence to the original learned distribution?   
\end{enumerate}

\vspace*{-5pt}
\paragraph{Problem instances and data.} We ran experiments on the following 3 problem instances. 
\vspace*{-5pt}
\begin{itemize}[topsep=5pt,leftmargin=*]
\item \texttt{Toy} was synthetically generated by randomly truncating edges on a layerwise-fully-connected layered graph. It features 11 layers and 41 vertices. 
\item \texttt{Heights} and \texttt{Ridge} are layered graphs converted from real-world city maps using the method of \citep{cerny2024layered}. 
\texttt{Heights} is a middle-scale graph with 12 layers and 423 vertices while \texttt{Ridge} is larger, with 13 layers and 4293 total vertices. These graphs are relatively sparse, 
with out-degree of each vertex mostly being less than 6.
\end{itemize}

\vspace*{-5pt}
The \texttt{Toy} dataset comprises all unique paths in the graph, while the dataset for \texttt{Heights} ($\sim$120k paths) and \texttt{Ridge} ($\sim$347k paths) were obtained by sampling them \textit{nonuniformly} over all paths. Note that in both cases, the number of paths are orders of magnitude higher. We employ UNet-like neural networks \citep{ronneberger2015u} as the backbone of the diffusion models, use 256 diffusion timesteps and train them with the AdamW optimizer~\citep{loshchilov2017decoupled}.
All experiments were run on 
an AMD Ryzen Threadripper PRO 5995WX 64-Core CPU, 504 GB RAM, and 2 NVIDIA RTX A6000 GPUs each with 48GB GPU memory using PyTorch \citep{paszke2019pytorch}.

    \subsection{Structured Discrete Representation for Generation Quality}%
    \label{sub:discrete_quality}

    \begin{table}[t]
            \centering
            \begin{minipage}{0.5\textwidth}
                \centering
                \begin{tabular}{lcc}
                {\textsc{Method}}  &{\bf Valid Rate (VR) \% $\boldsymbol{\uparrow}$}  \\ 
                \hline
                \edpgnn          & $0.00$\\
                \swingnn         & $99.7$ \\
                \method (Ours)   & {$\mathbf{100}$} \\
                \end{tabular}
                \caption{
                    Valid rate comparisons in unconditional path generation. 
                    Each method generates 8192 samples for validity check. 
                }
                \label{tab:vr_uncond}
            \end{minipage}
            \hfill
            \begin{minipage}{0.45\textwidth}
                \centering
                \includegraphics[width=0.7\linewidth]{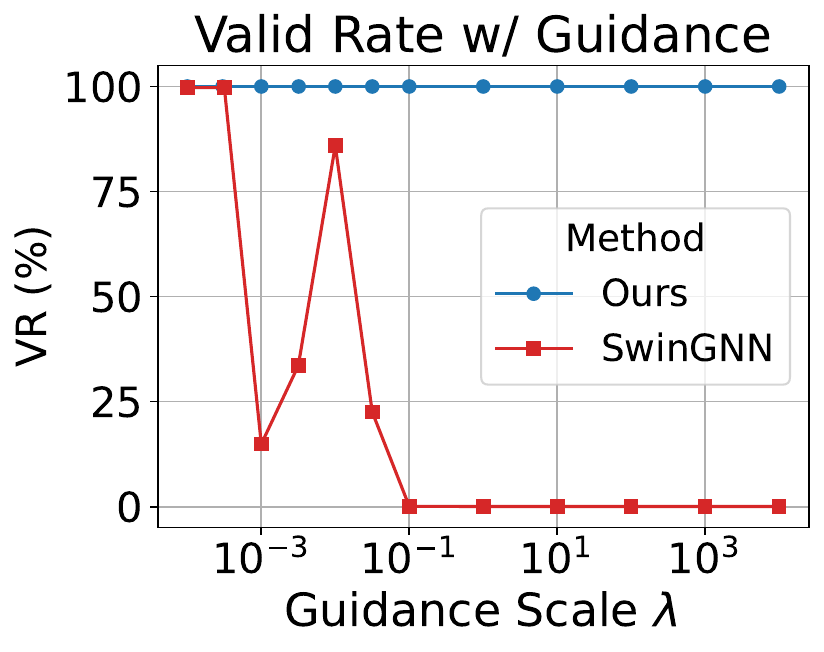} 
                \captionof{figure}{
                    Valid rates for baseline and our method under different guidance strengths. 
                }
                \label{fig:vr_cmp}
            \end{minipage}
        \vspace*{-15pt}
        \end{table}

    To address {\bf RQ1}, 
    we compared the sample quality of our proposed method to diffusion baselines that operate on the continuous relaxation of the discrete data domain. By sample quality, we refer to the proportion of \textit{valid paths} that are generated. Recall from \secref{sec:palm} that by construction, our proposed method generates valid paths with certainty.

    \paragraph{Unconditional path generation.} We compared our method to baselines \edpgnn~\citep{niu2020permutation} and \swingnn~\citep{yan2024swingnn}. 
    Both \edpgnn and \swingnn 
    generate adjacency matrices rather than paths.\footnote{
    \edpgnn learns distributions of adjacency matrices of graphs by training a graph neural network via score matching and sampling with annealed Langevin dynamics. 
    \swingnn improves upon \edpgnn by leveraging a graph transformer to learn the score of adjacency matrices distributions and perform sampling from the the perspective of diffusion stochastic differential equations.}
    Thus, there is non-zero probability where the edges included in the adjacency matrix do not form a single path in the layered graph.
    For each of the 3 methods, we trained a denoising diffusion model using the same training data, with the only difference in how paths were represented internally. For this part, no guidance was used, \ie, only unconditional sampling was tested.  
    To evaluate sample quality, we generated a set $S$ of $N$ samples and report the valid rate (VR):
        \begin{align}
        \label{eq:vr}
            \mathrm{VR} \triangleq 
                \frac{
                    \sum_{s\in S} \mathds{1}\left[ s \text{ is a path} \right]
                }{N} \times 100\%
        .\end{align}

    We used a dataset of size 1350 with at 80-20 train-validation split. The results are reported in Table~\ref{tab:vr_uncond}. As expected, our method yields a VR of 100\% since our approach by construction guarantees a valid path. \edpgnn yields a VR of $0$, meaning that essentially no valid paths were generated. Surprisingly, \swingnn generates valid paths almost all the time. 

    \paragraph{Conditional path generation with reward guidance.} To further investigate this phenomenon, we consider the more complex task of conditional path generation, \ie, diffusion \textit{with guidance} and compare our method to \swingnn for various levels of guidance scales $\lambda$. 
    We utilize the models trained earlier for sampling under reward predictor guidance as condition generation, following the general idea of classifier guidance~\citep{dhariwal2021diffusion}. 
    We adopt a simple reward setup: a path obtains $1.0$ reward if it contains a specific edge in the layered graph. 
    Since \swingnn does not originally support guidance, we adopt a simplified approach described by \citet{ma2024elucidating} 
    and leverage the gradients of rewards achieved by noisy samples as guidance signals. 
    Guided sampling with \method is implemented as described in \algoref{alg:ddps}. 
    \footnote{Since \swingnn uses an internal representation distinct from PALM, guidance scale technically cannot be compared across both methods.}
    
    The results for this setting are reported in Figure~\ref{fig:vr_cmp}. When $\lambda$ is small, both methods generated valid paths almost all the time, but as $\lambda$ increases, 
    \swingnn's performance drops (surprisingly, not monotonically in $\lambda$) to zero. In contrast, our method guarantees perfect valid rate by construction regardless of $\lambda$. These results point toward the brittle nature of continuous-state approximations for inherently discrete variables.

    \subsection{Discrete Posterior Sampling for Reward Improvement}%
    \label{sub:posterior_reward}
    \begin{figure}
        \centering
        \subfigure[\texttt{Toy}]{
            \includegraphics[width=.316\columnwidth]{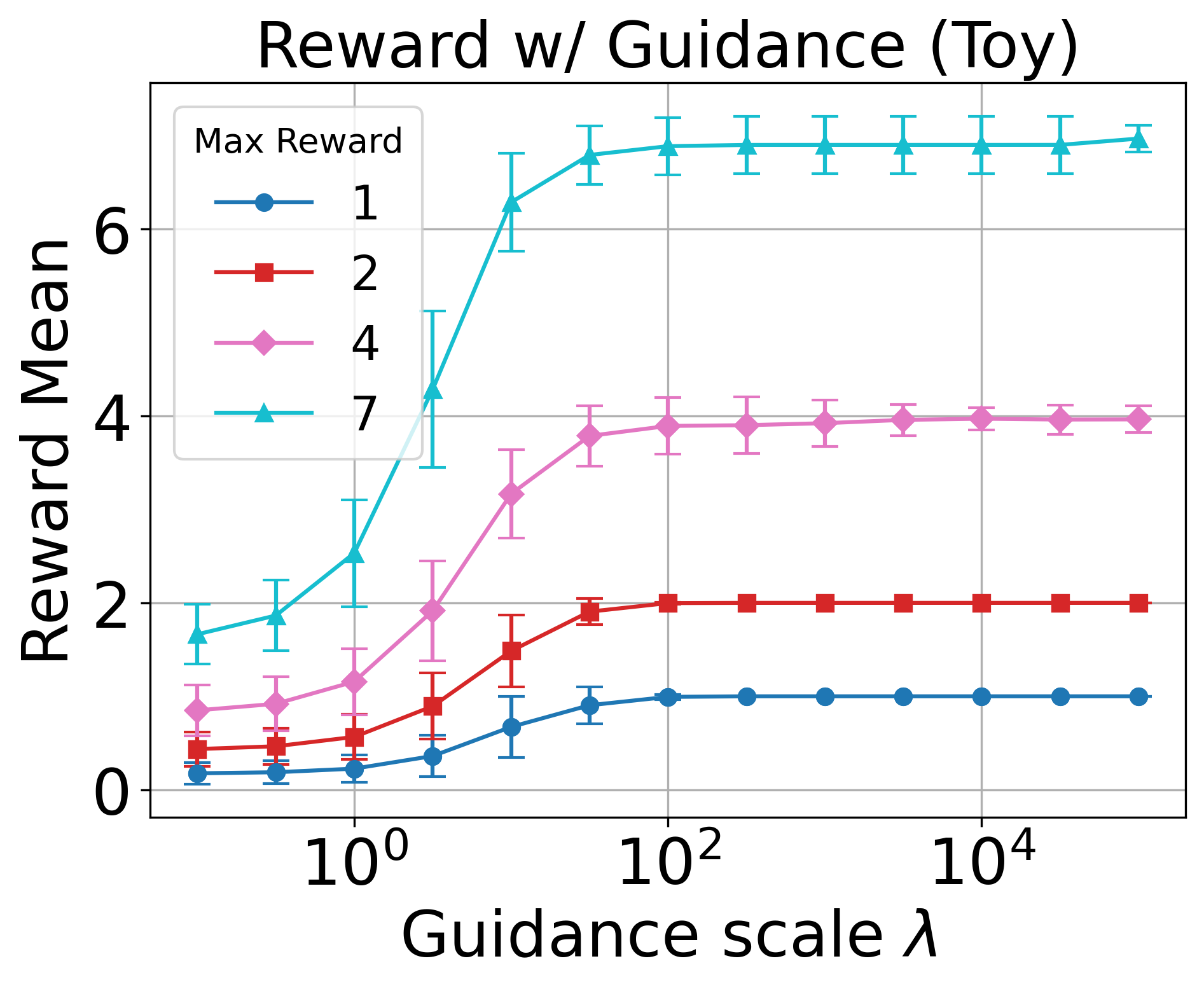}
            \label{fig:reward_fc}
        }
        \subfigure[\texttt{Heights}]{
            \includegraphics[width=.316\columnwidth]{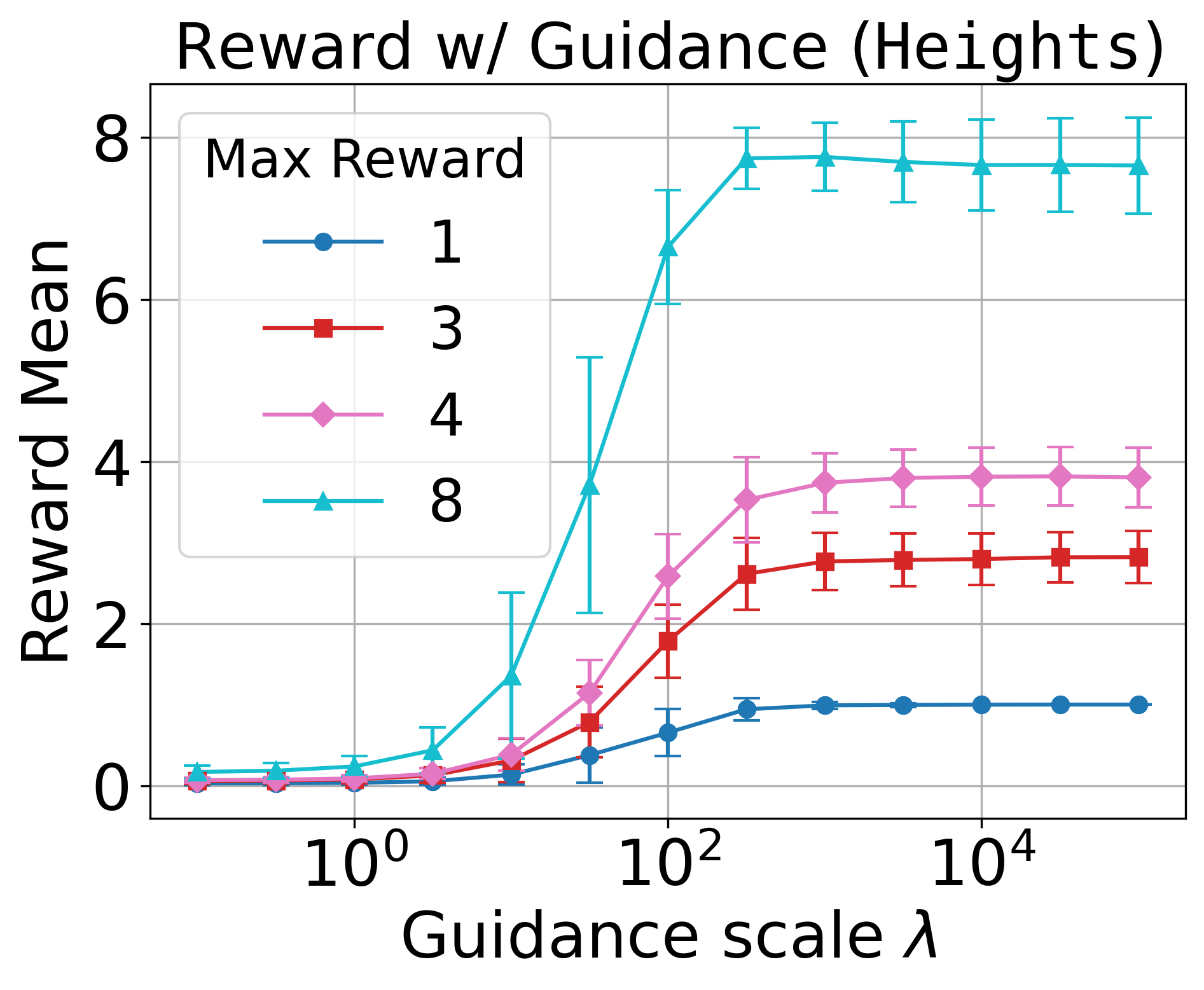}
            \label{fig:reward_columbia}
        }
        \subfigure[\texttt{Ridge}]{
            \includegraphics[width=.316\columnwidth]{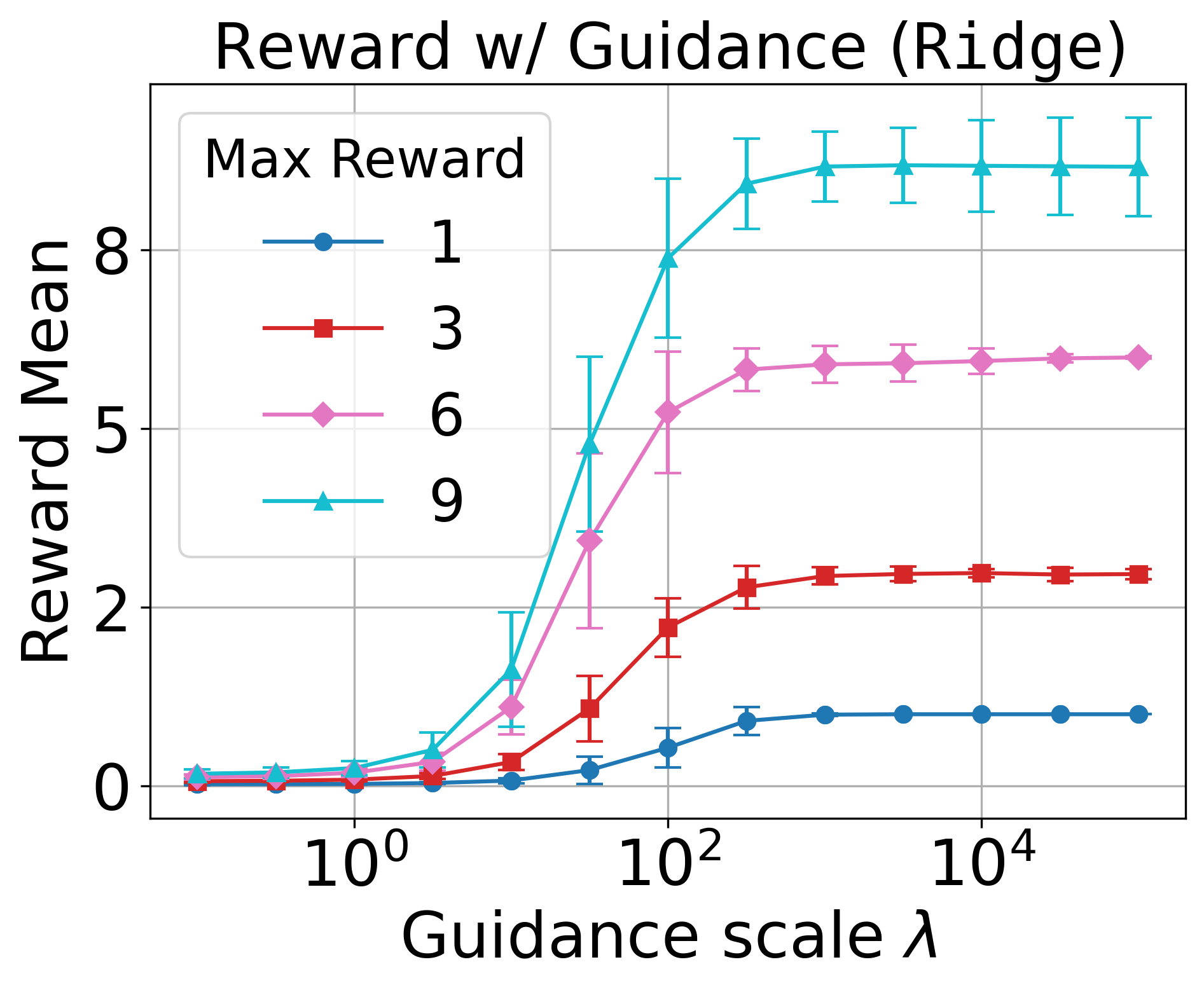}
            \label{fig:reward_nus}
        }
        \vspace*{-10pt}
        \caption{
            Average reward values under various guidance scales for different reward configurations. Error bars denote standard deviations. 
        }
        \label{fig:reward_improve}
    \vspace*{-10pt}
    \end{figure}

    To assess the ability of \method in reward improvement, we train diffusion models for all 3 problem instances, \texttt{Toy}, \texttt{Heights}, \texttt{Ridge}.\footnote{For \texttt{Toy}, the model was trained separately from \secref{sub:discrete_quality}.} 
    We follow the approach in \secref{sec:ddps} by assigning \textbf{binary} rewards to edges in a layered graph. An edge with a reward of $1$ indicates that paths containing it are favored, and \method should guide the sampling process into including these favored edges. We adopt 4 reward configurations, each specified by the maximum possible reward a path can ever achieve. For example, in \texttt{Toy}, these four configurations correspond to a maximum reward of 1,2,4, or 7 (\figref{fig:reward_fc}).
    For each problem instance, we took each configuration and ran guided sampling with different reward function instances, 20 for \texttt{Toy} and $\texttt{Heights}$, and, 10 for $\texttt{Ridge}$. 
    These classes of configurations characterize a range of reward sparsity--- from nearly half of all possible paths achieving maximum rewards to the case where a path chosen u.a.r. achieves it with odds poorer than one over ten million.

    For each layered graph, we used the trained discrete diffusion model as the base sampler, and perform guided sampling with \method. 
    We evaluate reward improvement by taking the empirical means of rewards with 65536 samples for \texttt{Toy}, 4096 for \texttt{Heights} and 2048 for \texttt{Ridge}. 

    We present in \figref{fig:reward_improve} the average rewards for different reward functions and under different guidance scales of \method. 
    The trends observed in \figref{fig:reward_improve} agree with what we expect: with stronger guidance, the higher reward one obtains eventually plateauing at the maximum obtainable reward. The ``S''-shaped curve (note the logarithmic x-axis) suggests that when $\lambda$ is small then the gradient of the obtained reward with respect to $\lambda$ is small as well.
    Overall, our results suggest that \method leads to reward improvement as long as a sufficiently large guidance scale is selected.

    \subsection{Trade-offs between Reward Objective and Learned Distribution}
    \label{sub:tradeoffs} 
    While rewards obtained are highest with a large enough guidance scale $\lambda$, this comes at the expense of deviating from the original (\ie, without guidance) distribution. We examine this tradeoff empirically by examining path distributions as $\lambda$ varies. 
    To do so, we introduce the idea of a \textit{target distribution}. A desirable target distribution in our case for reward-guided sampling would be the posterior distribution of paths given the true likelihood of our desired property (\ie, obtaining max rewards) and the learned prior (without guidance). 
    In our context, we look at the distribution of paths when $\lambda=0$, \textit{conditioned on them receiving the maximum achievable reward} (e.g., 1,2,4,7 for \texttt{Toy} in \figref{fig:reward_fc}) and compare that to the distribution of paths obtained when $\lambda > 0$. 
    
    We appeal to the following metrics between probability distributions to show the gap between the target and generation distributions $\ptar{x}$ and $\pgen{x}$, 
    \vspace*{-3pt}
    \begin{itemize}[topsep=3pt, partopsep=0pt, itemsep=1pt, parsep=1pt,leftmargin=*]
        \item KL divergence $\kldiv{\ptar{x}}{\pgen{x}}$; 
        \item $L^1$ distance: $d_{L^1}[{\ptar{x}}, {\pgen{x}}] = \sum_x \left| \ptar{x} - \pgen{x} \right|$; 
        \item Total Variation distance: $\tvd{\ptar{x}}{\pgen{x}} = \sup_x \left| \ptar{x} - \pgen{x} \right|$; 
        \item 
            Spearman's Footrule Distance: $\sfdord{\pi}{\ptar{x}}{\pgen{x}}$. 
            See \dfnref{dfn:sfd} in Appendix~\ref{apdxsub:metrics} for a formal definition.
            This metric covers the \textit{relative significance} of elements in terms of probability. 
    \end{itemize}
    However, as the size of layered graph increases, the number of total possible paths and the support of path distributions grows exponentially. 
    This makes it computationally intractable to estimate $\ptar{x}$ and $\pgen{x}$ by sampling. 
    Therefore, for such large graphs we evaluate distributions of certain \textit{features} of the sampled paths, where these features are much lower-dimensional:
    \begin{itemize}[topsep=2pt, partopsep=0pt, itemsep=1pt, parsep=1pt,leftmargin=*]
    \item \gls{flgd} is an adaptation of the \gls{fid} in image generation. 
    The feature vector of a sample path is obtained by flattening the \gls{palm} representation of the sample and zeroing out all entries belonging to vertices outside the path. 
    \item \Gls{isl} aggregates across layers the marginal distributions over vertices in the each layer. We employ four variants based on the measure: \isllone, \islkl, \isltv, and \islsf, each corresponding to the metrics used on path distributions as discussed above. 
    A formal definition of \gls{isl} is in \dfnref{dfn:isl} in Appendix~\ref{apdxsub:metrics}. 
    \end{itemize}

    \begin{figure}
    \centering
    \subfigure[$d_{L^1}$ $\boldsymbol{\downarrow}$]{
        \includegraphics[width=.232\columnwidth]{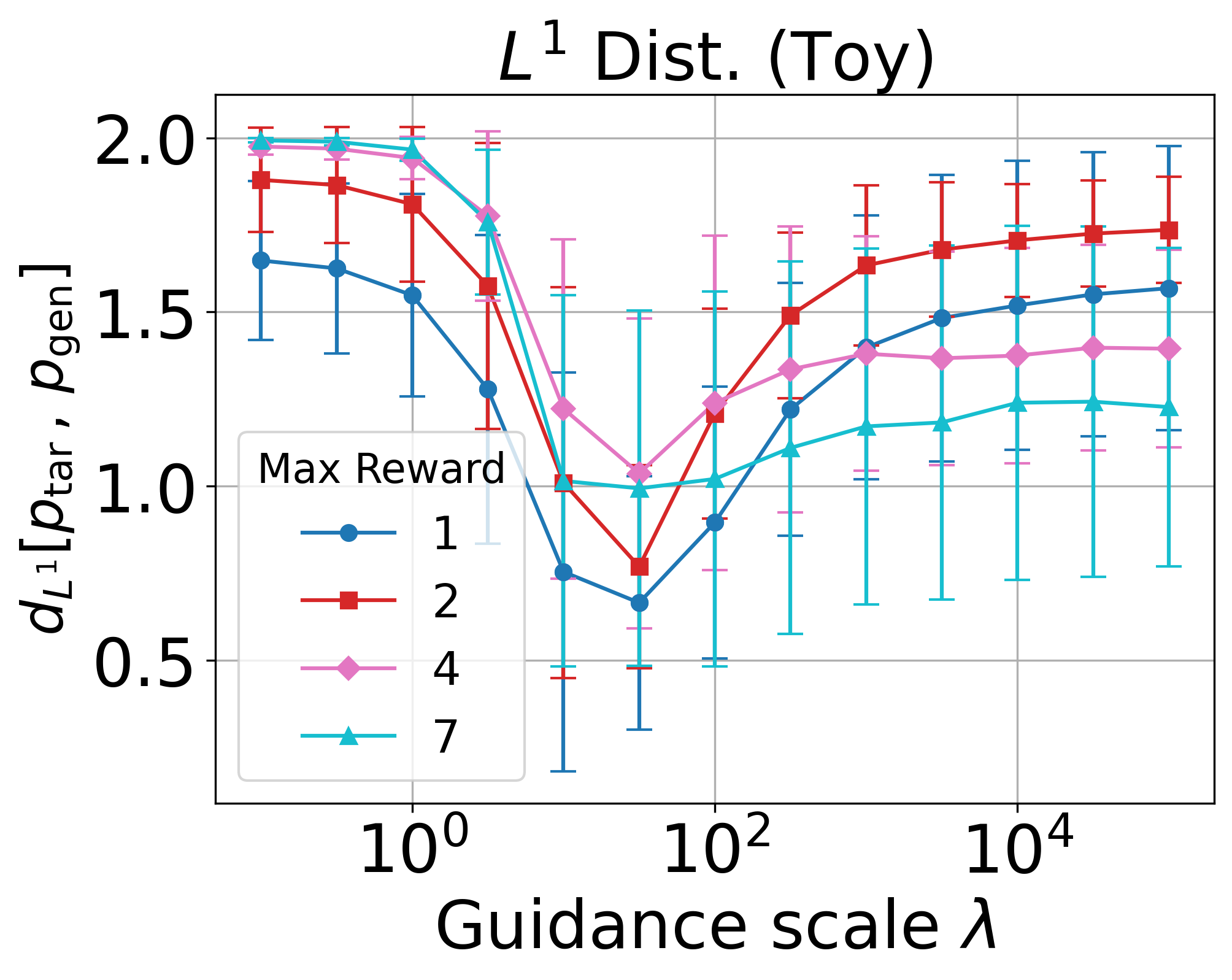}
        \label{fig:l1_fc}
    }
    \subfigure[$d_{\mathrm{TV}}$ $\boldsymbol{\downarrow}$]{
        \includegraphics[width=.232\columnwidth]{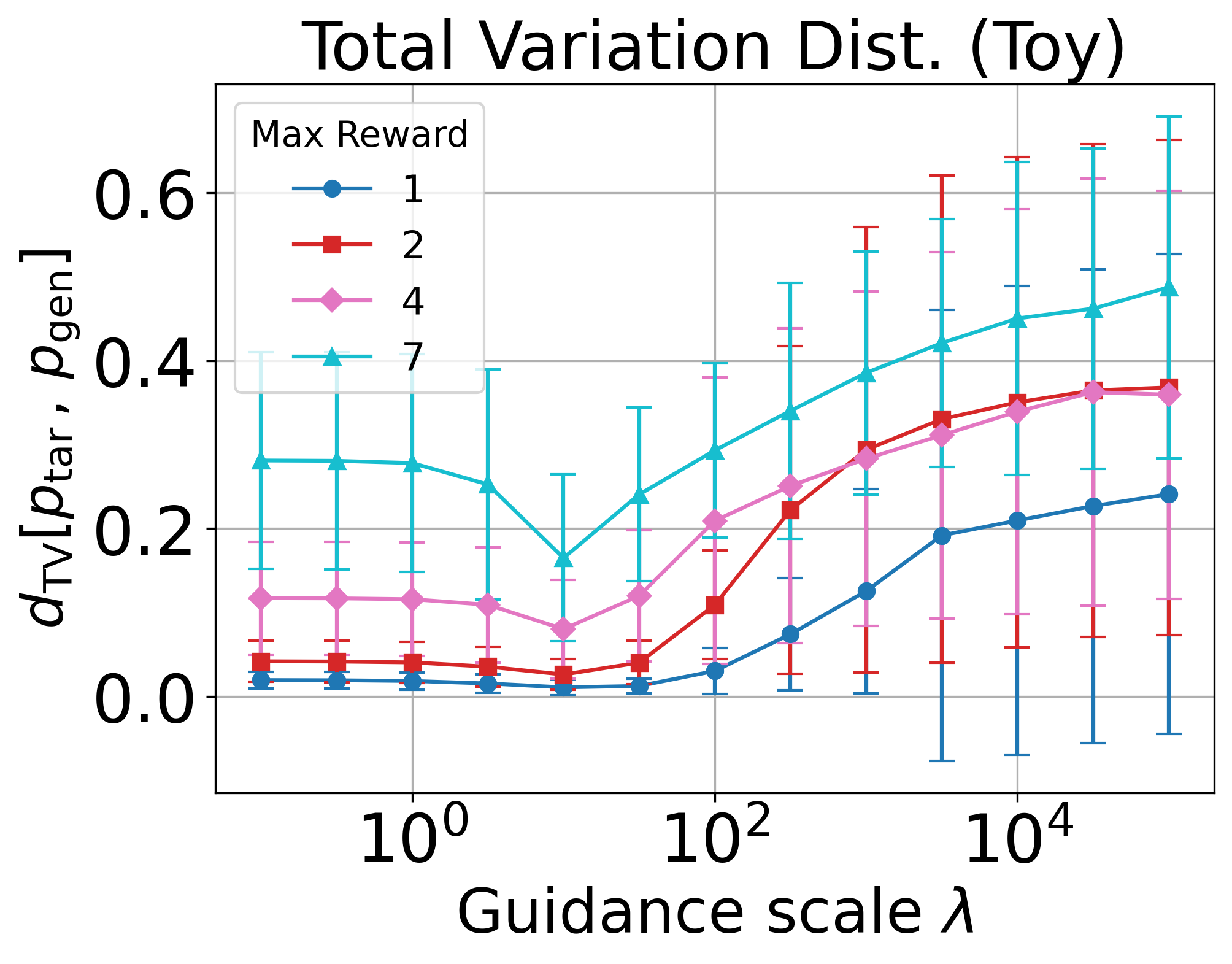}
        \label{fig:tvd_fc}
    }
    \subfigure[SFD $\boldsymbol{\downarrow}$]{
        \includegraphics[width=.232\columnwidth]{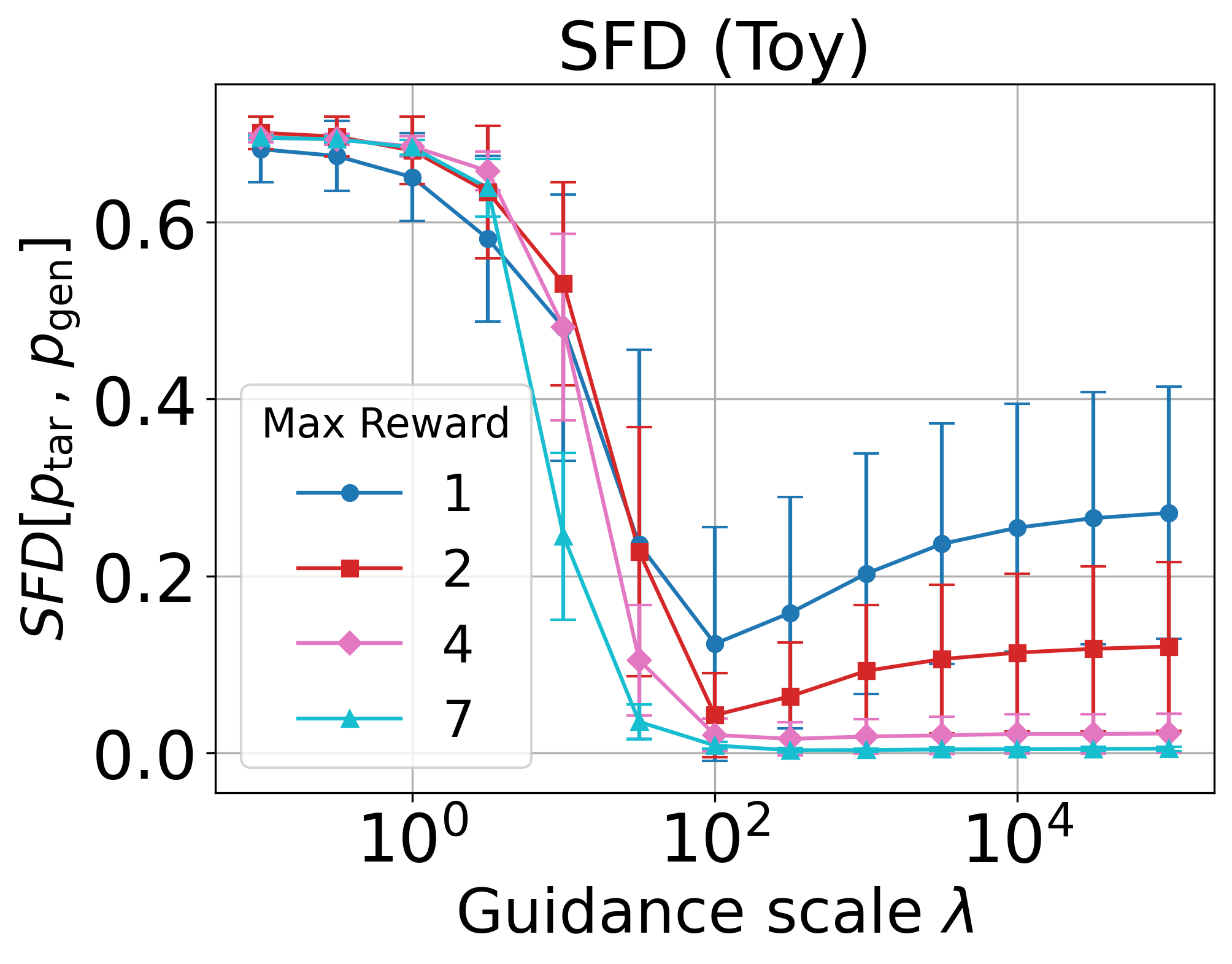}
        \label{fig:sfd_fc}
    }
    \subfigure[KL $\boldsymbol{\downarrow}$]{
        \includegraphics[width=.232\columnwidth]{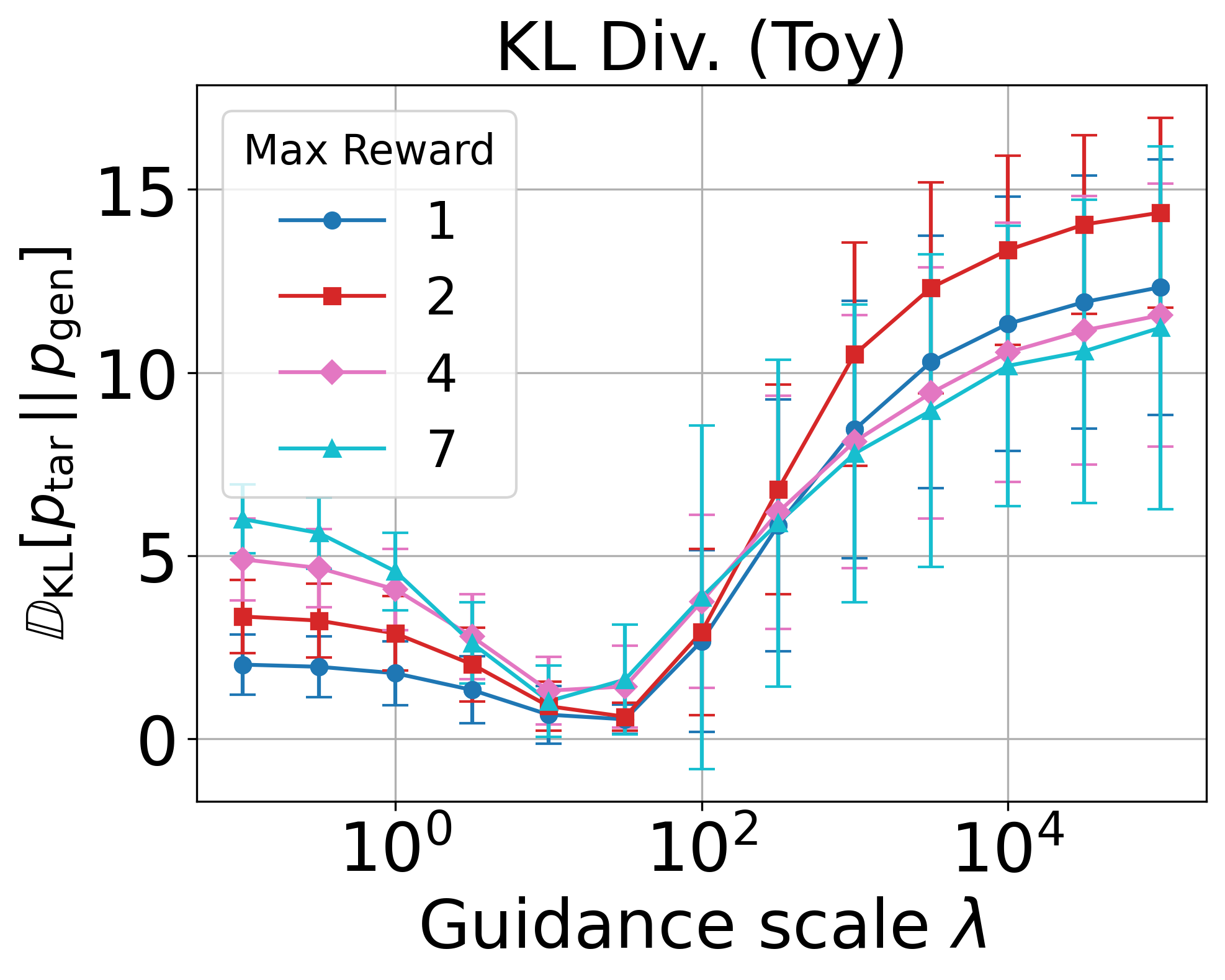}
        \label{fig:kl_fc}
    }
    \vspace*{-10pt}
    \caption{
        Metrics in path distributions for the small-size graph \texttt{Toy}. 
        The error bars represent standard deviations across different reward function instances within the reward configuration class (20 instances per configuration). 
        The sample size is 65536. 
    }
    \label{fig:dist_metrics_fc}
    \vspace*{-15pt}
    \end{figure}

    \Figref{fig:dist_metrics_fc} shows our results for \texttt{Toy} for all 4 metrics. The same trend is observed: as $\lambda$ increases, we see the distributions getting closer. This trend decreases until around $\lambda=100$, after which distances rise and eventually plateau.
    The shapes of the curves suggest a ``sweet spot'' balancing the reward objective and adherence to the learned prior distribution. 

    \Figref{fig:scores_all} in Appendix~\ref{apdx:result} shows our results using feature distributions for \texttt{Toy}, \texttt{Heights}, and \texttt{Ridge}. 
    By and large, they reveal a similar pattern--- except for scores calculated using the SFD, the scores decay as the $\lambda$ increases and achieve a ``sweet spot'' before rising and eventually plateauing. 
    With the exception of \islsf, the ``sweet spots'' are marked by the golden dashed lines, whose location approximately agree. This value of $\lambda$ achieves near-optimal reward. This suggests that with the right amount of guidance, \method achieves a good tradeoff between reward and adherence to the original distribution (conditioned on attaining optimal reward).

\section{Conclusion}
In this paper we propose \method, which utilizes the \gls{palm} representation to guarantee that samples from a discrete diffusion model do indeed correspond to paths in a layered graph. Our preliminary results look favorable and we show that with classifier guidance, we are able to achieve high rewards given some reward function while still maintaining reasonable adherence to the learned distribution. Future work includes extending this to more general problems beyond path generation, or involving more elaborate constraints or rewards on the generated paths.

%% file: sections/appendices.tex
\appendix

\section{Acknowledgments}

We acknowledge the usage of Claude \citep{claude} to generate the term and acronym ``padded adjacency-list matrix (PALM).'' 
In addition, we used Claude \citep{claude} to polish a few paragraphs in \secref{sec:ddps} and \secref{apdxsub:metrics}.

\input{sections/app_C_expt_details}

\input{sections/app_D_results}

%% file: sections/app_C_expt_details.tex
\section{Experiment Details}
\label{apdx:expt}

\subsection{Data}
    Datasets used in \secref{sec:experiments} was generated synthetically. 
    The two small layered graphs (one used in \secref{sub:discrete_quality} and \texttt{Toy}) were obtained by first construct a layerwise-fully-connected graphs and then pruning 50\% of all the edges within. 
    The pruning process is to choose edges in the graph uniformly at random, and then remove it. 
    Checks were performed to ensure the resulting pruned graph meets definition of layered graphs and there exists paths from the first layer to the last layer. 
    All possible unique paths in the two pruned graphs were included in the datasets. 
    For \texttt{Heights} and \texttt{Ridge}, we first obtain all possible unique paths in the layered graphs by searching, and then sample paths from all possible paths uniformly at random without replacement. 
    After that, we duplicate the chosen paths by different numbers for each to create the nonuniform pattern in the dataset.

\subsection{Diffusion Model Implementation Details}

    We provide implementation details regarding training and inference of discrete diffusion models with \gls{palm}.

    \paragraph{Transition matrix construction}
        For vertices $v$ with positive out-degree(s) ($D_v > 0$): 
        \begin{equation}
            [Q^v_t]_{ij} = 
            \begin{cases}
                1 - \dfrac{D_v-1}{D_v}\beta_t, &\text{if}\quad 1\le i=j \le D_v \\
                \dfrac{1}{D_v} \beta_t,        &\text{if}\quad 1\le i\ne j \le D_v \\
                0,                             &\text{otherwise}.
            \end{cases}
        \end{equation} 
        For vertices with zero out-degree ($D_v = 0$):
        \begin{equation}
            Q^v_t = \I_D.
        \end{equation} 
        Note that for each transition matrix $Q^v_t$, its upper-left $D_v\times D_v$ block is a doubly stochastic square matrix.

    \paragraph{Inference with \gls{palm}}
        The only difference between the inference process of \method and that of \dddpm lies in that for each node $v$, we only take the first $D_v$ logits (corresponding to the $D_v$ edges originating at $v$) to construct the categorical distribution for sampling at each time step, since other entries are merely paddings and do not represent valid edges in the graph.

\subsection{Metrics}
\label{apdxsub:metrics}

We define here some metrics we used in the experiments in \secref{sub:tradeoffs}. 

\begin{defn}[Layer Imitation Scores (\isl)]
    \label{dfn:isl}
        Let $p\left( v^{(l)} \right),\, \forall l\in \{1,\ldots,L\}$ be the marginal distribution of vertices within layer $l$ of a layered graph with $L$ layers in total. 
        Given a path $x$, let the likelihood of vertices be the Dirac delta on the vertex on this path $x^{(l)}$ for each layer: 
        \begin{equation}
            \condpdf{p}{v^{(l)}}{x} = \delta \left( v^{(l)} - x^{(l)} \right) \, , \quad \forall l=1,\ldots,L 
        .\end{equation}
        Then, for a statistical dissimilarity measure $d[p, q]\,:\, \mathcal{P}\times\mathcal{P} \to \R$ between two discrete probability distributions, the \gls{isl} is defined as 
        \begin{equation}
            \operatorname{IS-L} \triangleq 
            \sum_{l=1}^{L} d\left[
                \myeexpect{w\sim \ptar{x}}{\condpdf{p}{v^{(l)}}{w}} ,\,
                \myeexpect{w^\prime\sim \pgen{x}}{\condpdf{p}{v^{(l)}}{w^\prime}} 
            \right]
        .\end{equation}
    \end{defn}

    \begin{defn}[Spearman's Footrule Distance (SFD), adapted from \citep{diaconis1977spearman}
    \footnote{
        The SFD defined in \dfnref{dfn:sfd} is adapted from, but slightly different to the Spearman's footrule in \citep[Eq.(1.1)]{diaconis1977spearman} which is defined over two non-parametric permutations. 
        In contrast, the two permutations involved in the SFD herein are parameterized by two probability distributions. 
    }]%
    \label{dfn:sfd}
        Given two discrete probability distributions $p$ and $q$ over a finite support $S = \{1, \ldots, n\}$, let $\pi_p$ and $\pi_q$ be the permutations of $S$ induced by sorting $p$ and $q$ in descending (or ascending) order by probability mass, respectively. 
        Specifically:
        \begin{itemize}
            \item 
                $\pi_p(i)$ represents the position of support element $i$ in the sorted ordering of $p$; 
            \item 
                $\pi_q(i)$ represents the position of support element $i$ in the sorted ordering of $q$. 
        \end{itemize}
        The Spearman's footrule distance between distributions $p$ and $q$ is then defined as:
        \begin{equation}%
        \label{eq:sfd}
            \sfdord{\pi}{p}{q} \triangleq 
            \frac{1}{M(n)}
            \sum_{i \in S} \left| \pi_p(i) - \pi_q(i) \right| 
        ,\end{equation}
        where $M(n)$ is the maximal possible distance, which occurs when one permutation is the reverse of the other:
        \begin{equation*}
            M(n) = \begin{cases}
            \begin{aligned}
                \textstyle \frac{n^2}{2} \, , &\quad \text{if } n \text{ is even}\\
                \textstyle \frac{n^2 - 1}{2} \, ,&\quad \text{if } n \text{ is odd}
            \end{aligned}
            \end{cases}
        \end{equation*}
    \end{defn}

    \begin{rmk}
    We acknowledge the SFD is \textit{not} invariant to permutations of elements with same probabilities. 
    As such, we utilize the same tie-breaking mechanism across all experiments to make fair comparisons. 
    \end{rmk}

%% file: sections/app_D_results.tex
\section{Additional Experiment Results}
\label{apdx:result}

    We include experiment results involving all feature-based metrics in this section. 
    
    \begin{figure}
    \centering
    \subfigure[Reward $\boldsymbol{\uparrow}$ \texttt{Toy}]{
        \includegraphics[width=.3\columnwidth]{figures/reward_fc.png}
        \label{fig:rwd_fc}
    }
    \hfill
    \subfigure[Reward $\boldsymbol{\uparrow}$ \texttt{Heights}]{
        \includegraphics[width=.3\columnwidth]{figures/reward_columbia.png}
        \label{fig:rwd_columbia}
    }
    \hfill
    \subfigure[Reward $\boldsymbol{\uparrow}$ \texttt{Ridge}]{
        \includegraphics[width=.3\columnwidth]{figures/reward_nus.png}
        \label{fig:rwd_nus}
    }
    \hfill
    \subfigure[IS-L $d_{L^1}$ $\boldsymbol{\downarrow}$]{
        \includegraphics[width=.3\columnwidth]{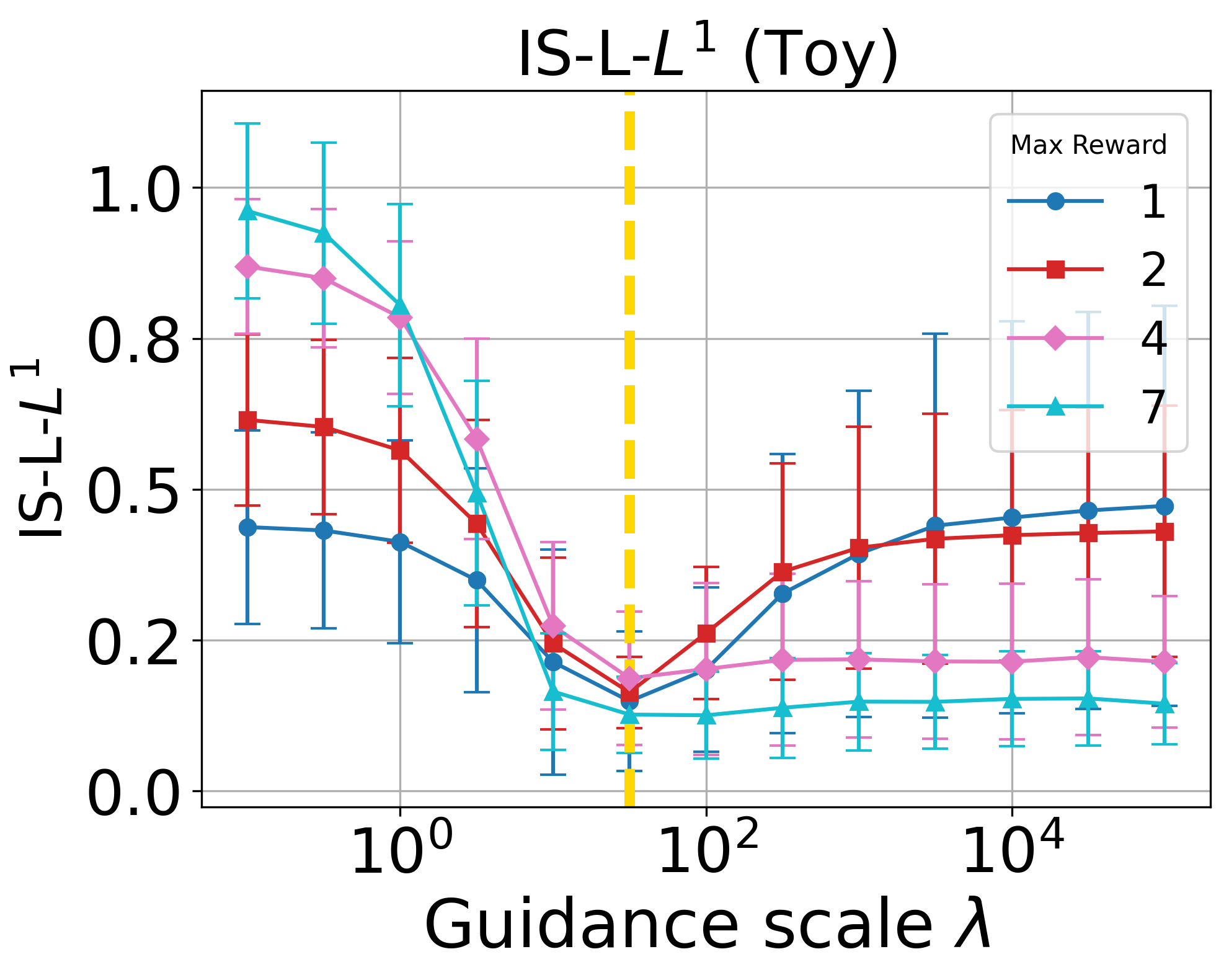}
        \label{fig:isll1_fc}
    }
    \hfill
    \subfigure[IS-L $d_{L^1}$ $\boldsymbol{\downarrow}$ \texttt{Heights}]{
        \includegraphics[width=.3\columnwidth]{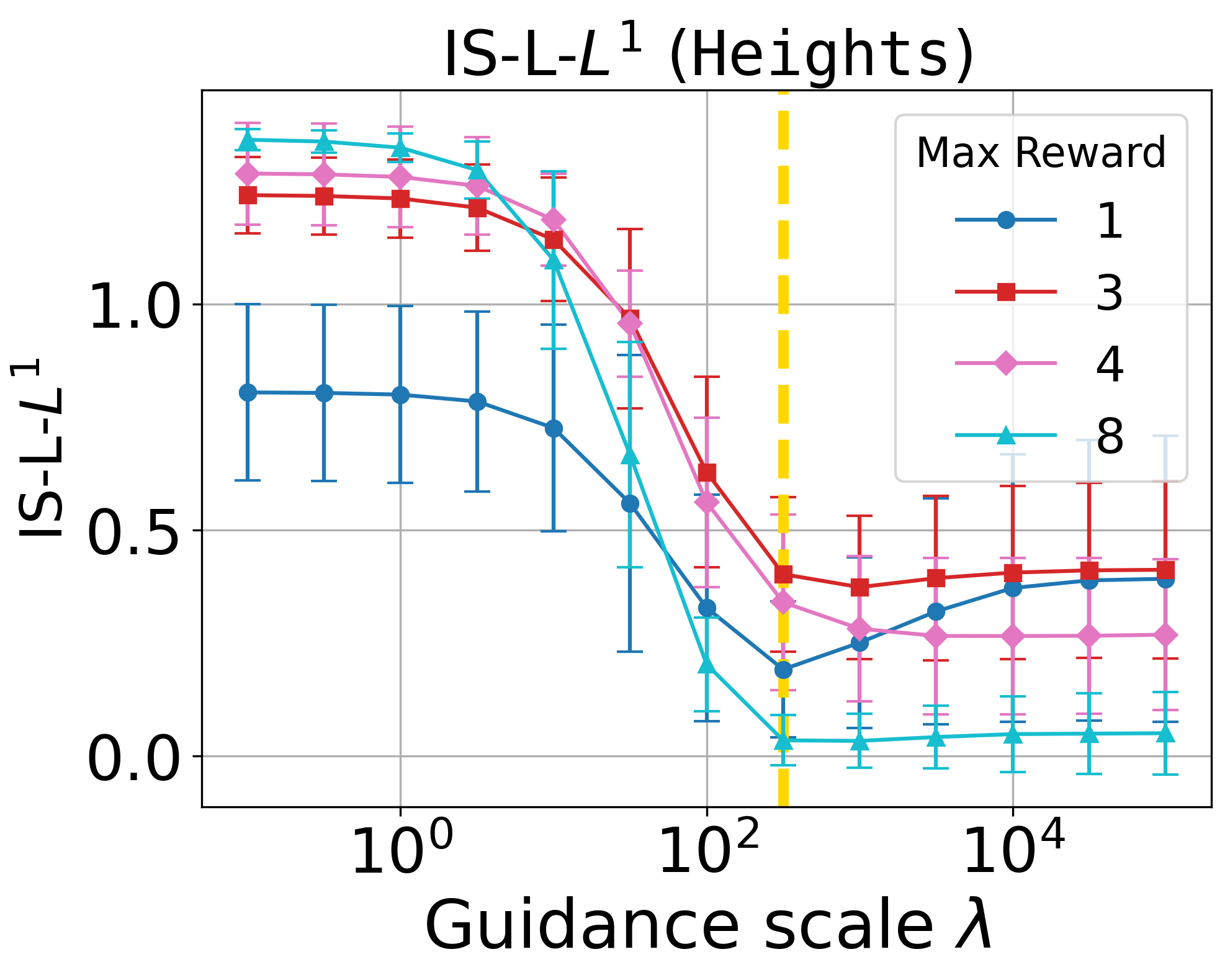}
        \label{fig:isll1_columbia}
    }
    \hfill
    \subfigure[IS-L $d_{L^1}$ $\boldsymbol{\downarrow}$ \texttt{Ridge}]{
        \includegraphics[width=.3\columnwidth]{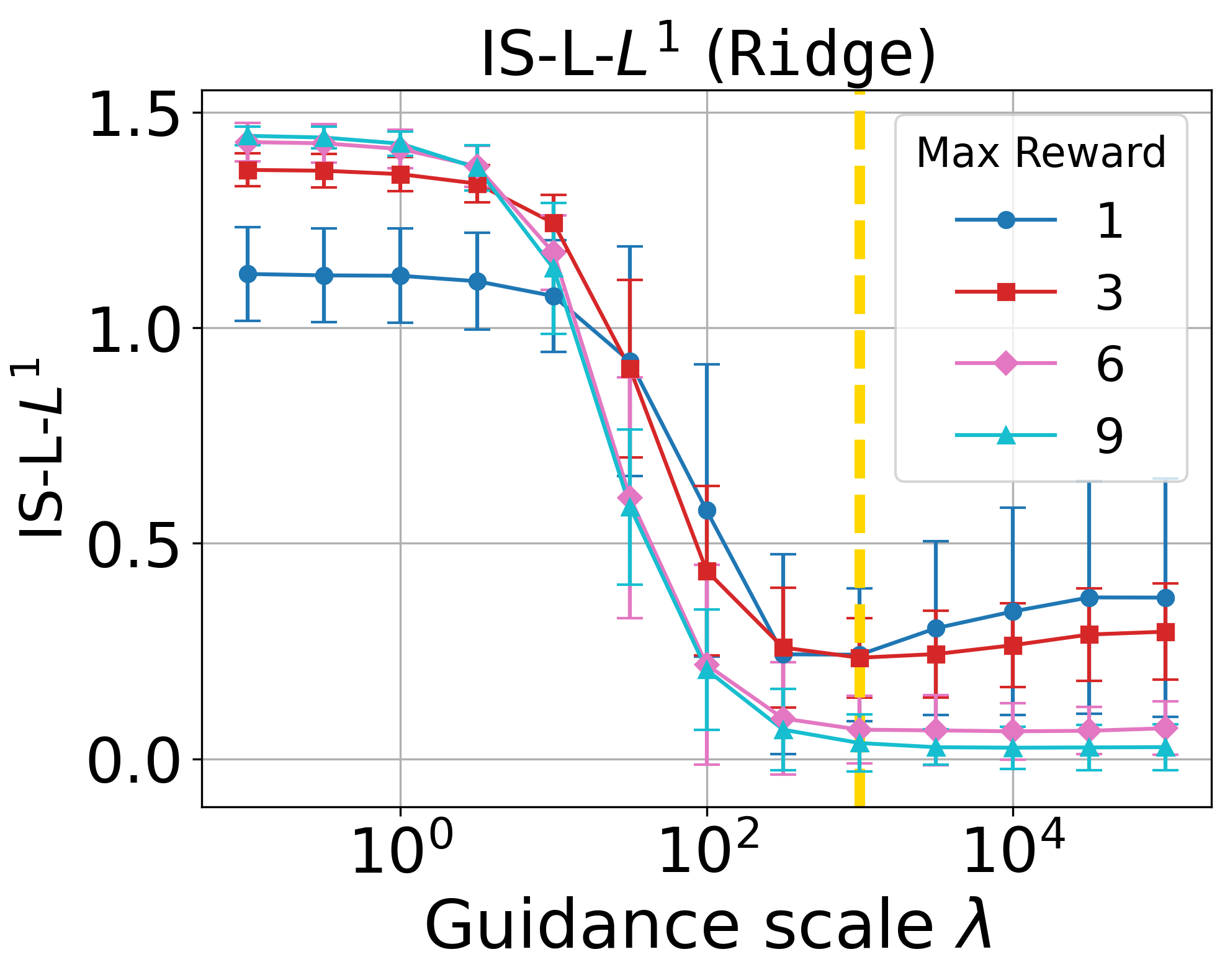}
        \label{fig:isll1_nus}
    }
    \hfill
    \subfigure[IS-L $d_{\mathrm{TV}}$ $\boldsymbol{\downarrow}$ ]{
        \includegraphics[width=.3\columnwidth]{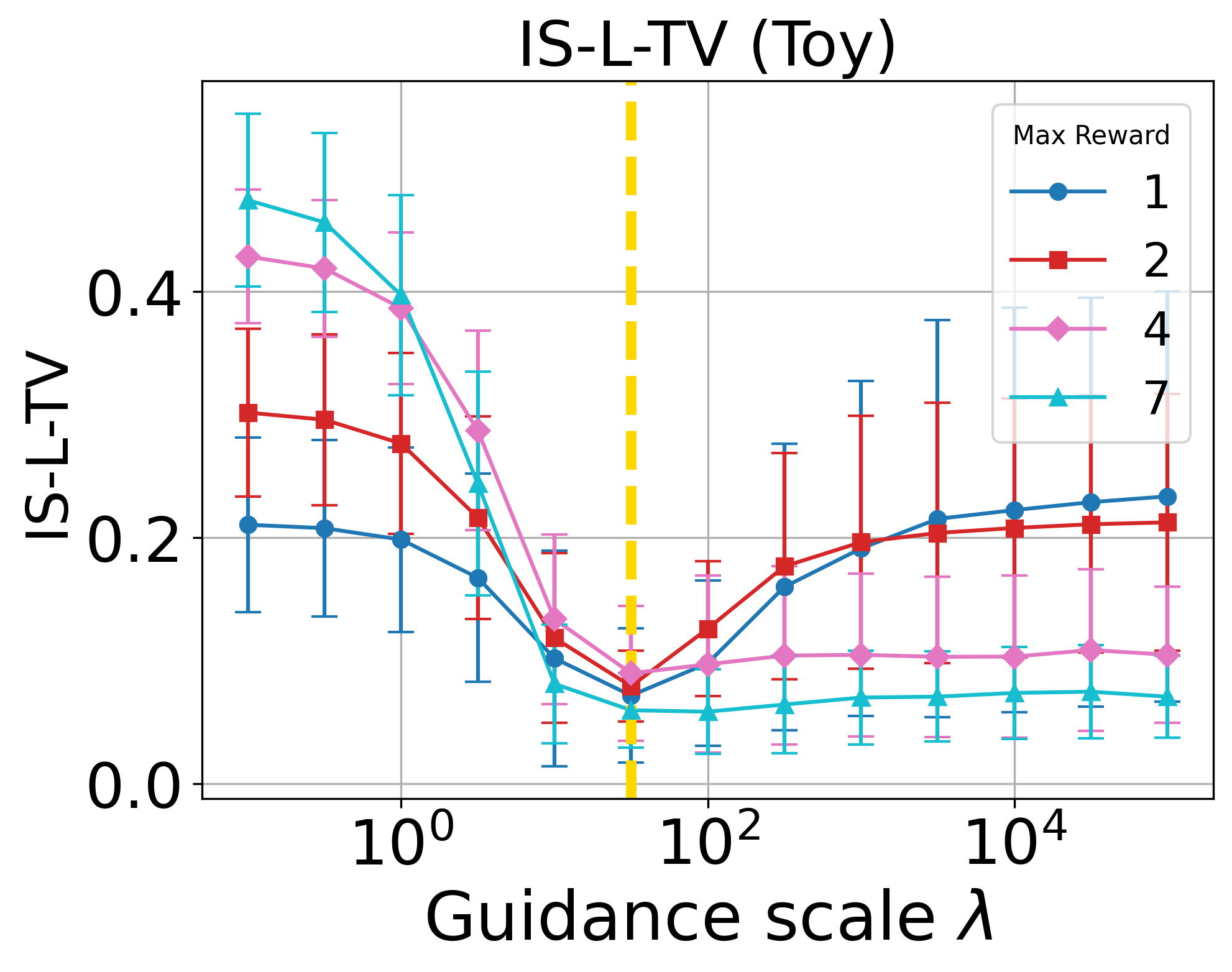}
        \label{fig:isltv_fc}
    }
    \hfill
    \subfigure[IS-L $d_{\mathrm{TV}}$ $\boldsymbol{\downarrow}$ \texttt{Heights}]{
        \includegraphics[width=.3\columnwidth]{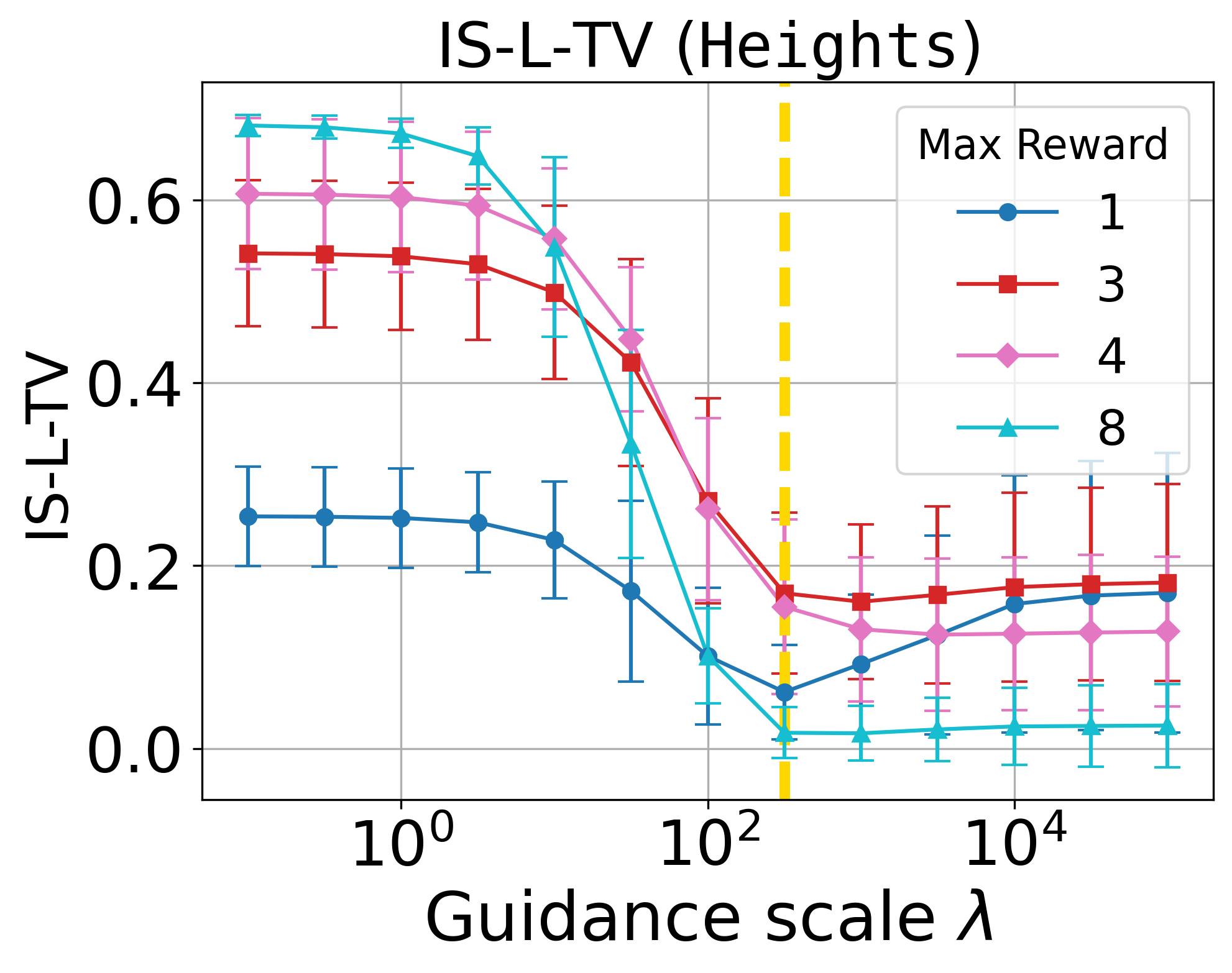}
        \label{fig:isltv_columbia}
    }
    \hfill
    \subfigure[IS-L $d_{\mathrm{TV}}$ $\boldsymbol{\downarrow}$ \texttt{Ridge}]{
        \includegraphics[width=.3\columnwidth]{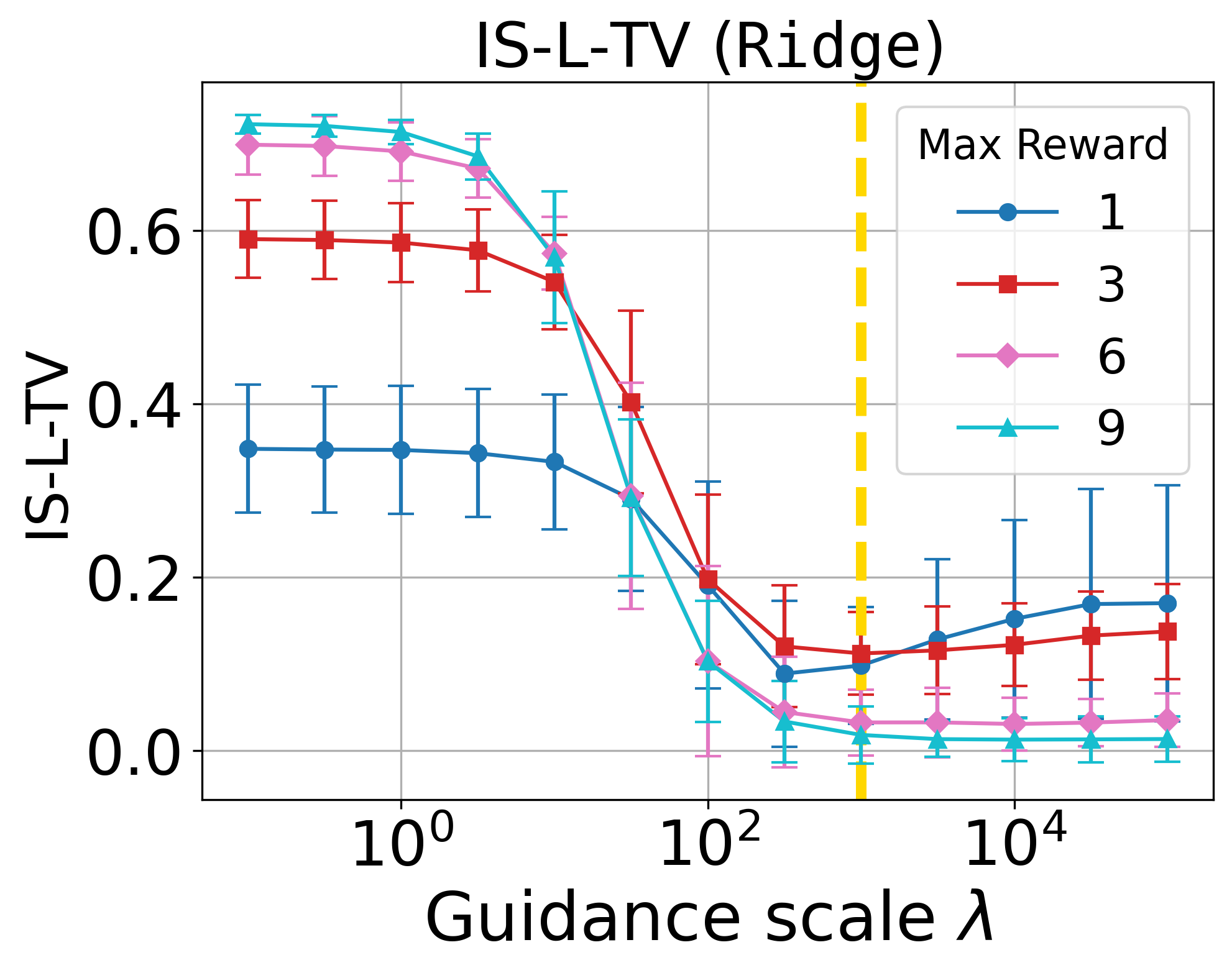}
        \label{fig:isltv_nus}
    }
    \hfill
    \subfigure[IS-L SFD $\boldsymbol{\downarrow}$]{
        \includegraphics[width=.3\columnwidth]{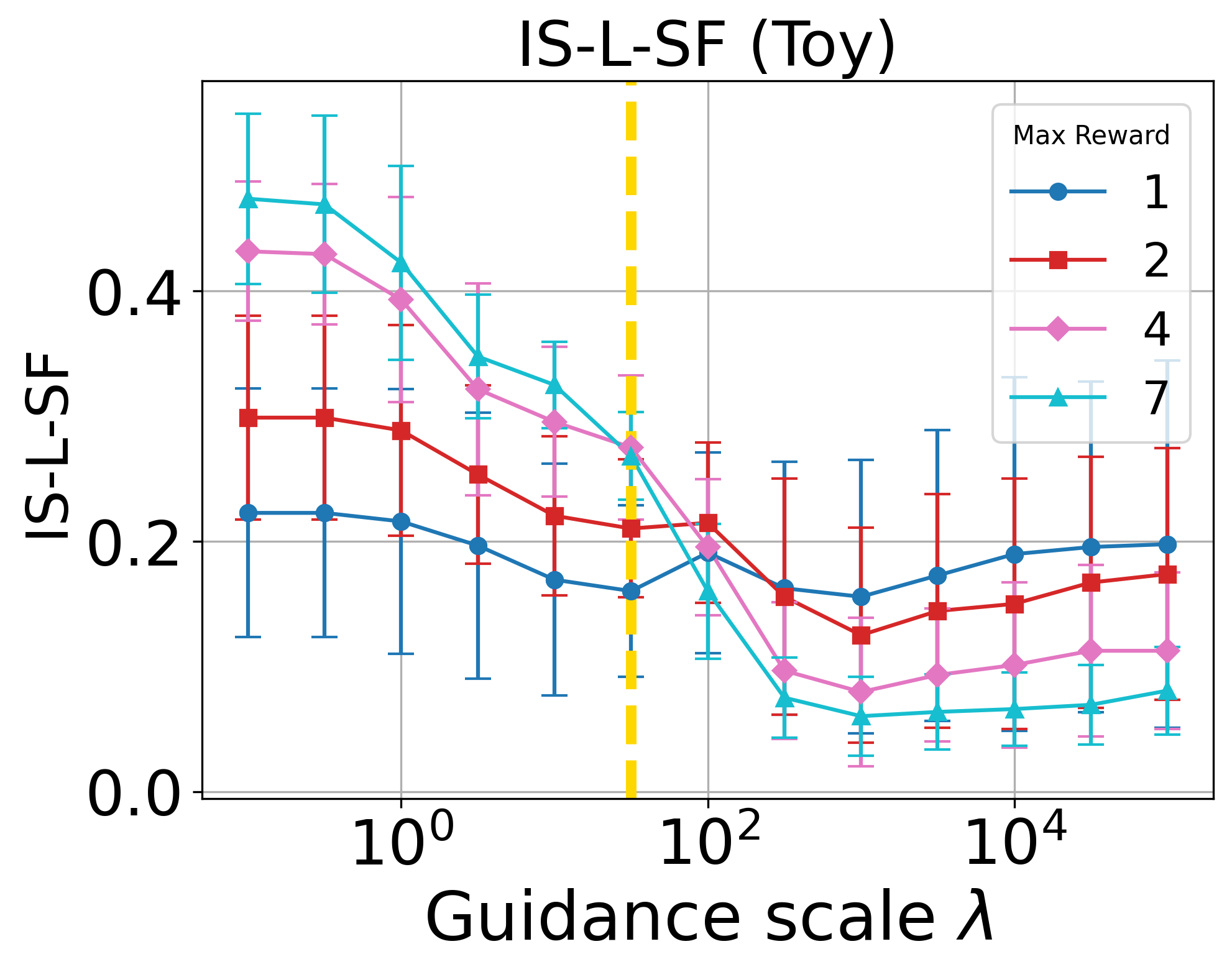}
        \label{fig:islsf_fc}
    }
    \hfill
    \subfigure[IS-L SFD $\boldsymbol{\downarrow}$ \texttt{Heights}]{
        \includegraphics[width=.3\columnwidth]{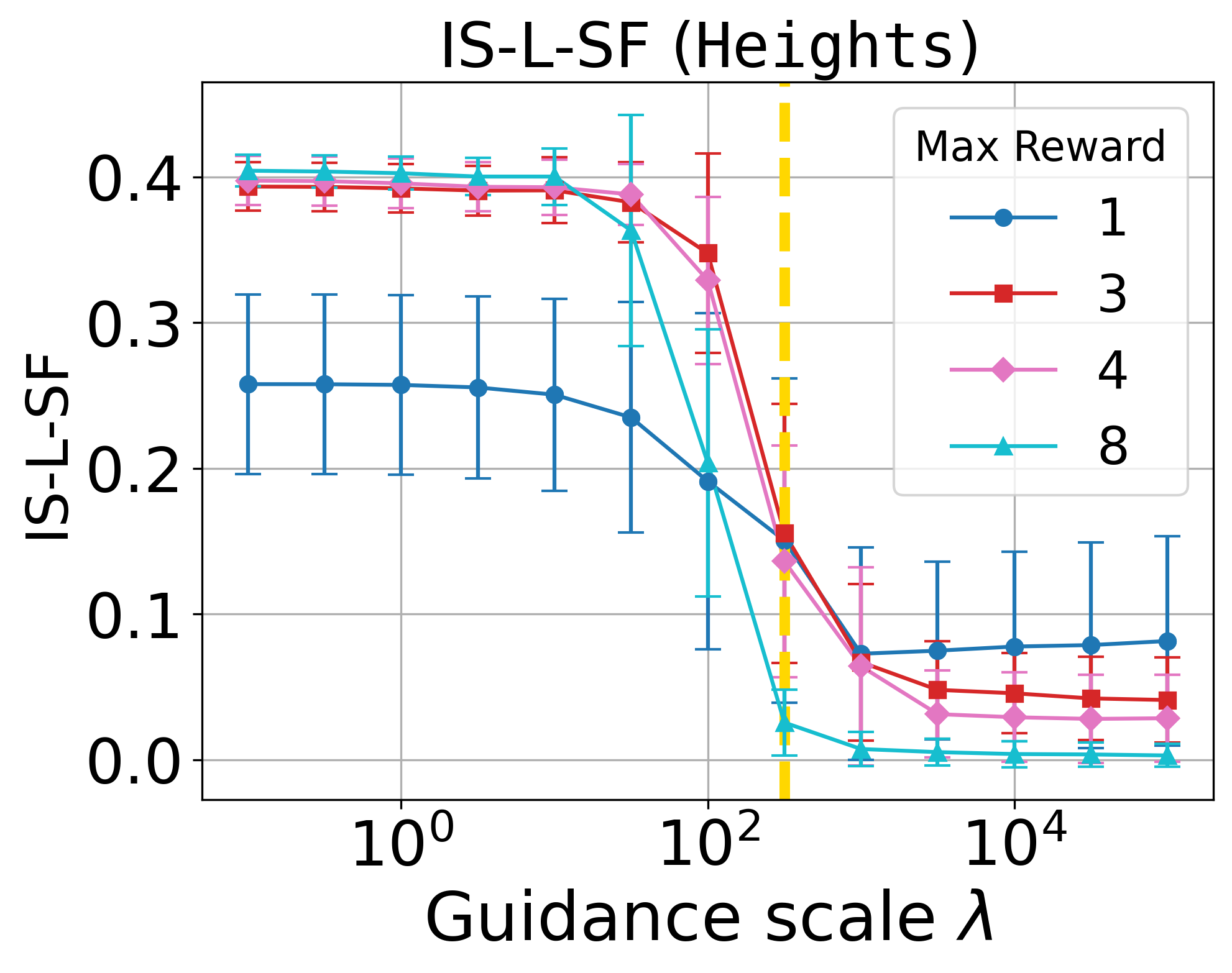}
        \label{fig:islsf_columbia}
    }
    \hfill
    \subfigure[IS-L SFD $\boldsymbol{\downarrow}$ \texttt{Ridge}]{
        \includegraphics[width=.3\columnwidth]{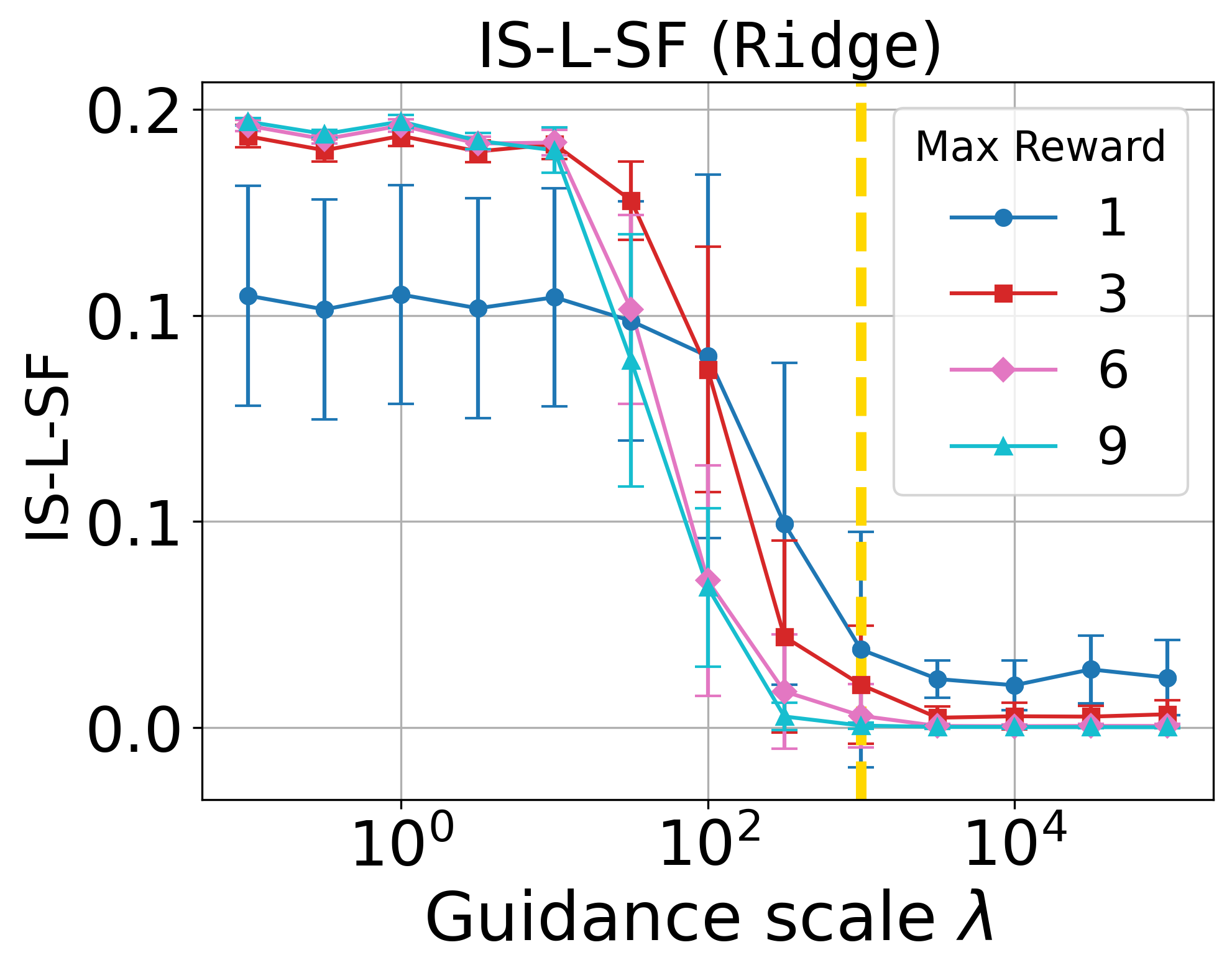}
        \label{fig:islsf_nus}
    }
    \hfill
    \subfigure[FLGD $\boldsymbol{\downarrow}$ \texttt{Toy}]{
        \includegraphics[width=.3\columnwidth]{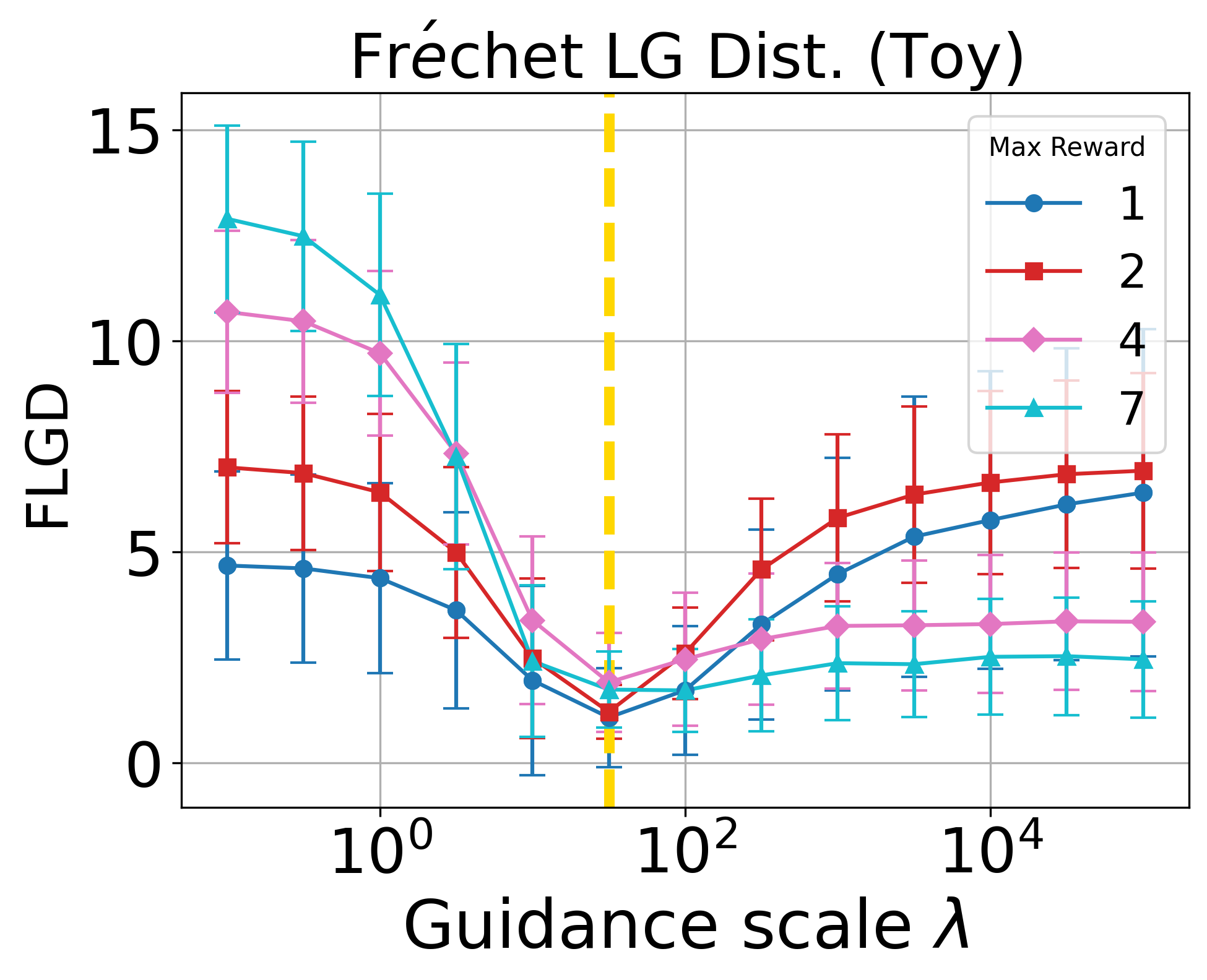}
        \label{fig:flgd_fc}
    }
    \hfill
    \subfigure[FLGD $\boldsymbol{\downarrow}$ \texttt{Heights}]{
        \includegraphics[width=.3\columnwidth]{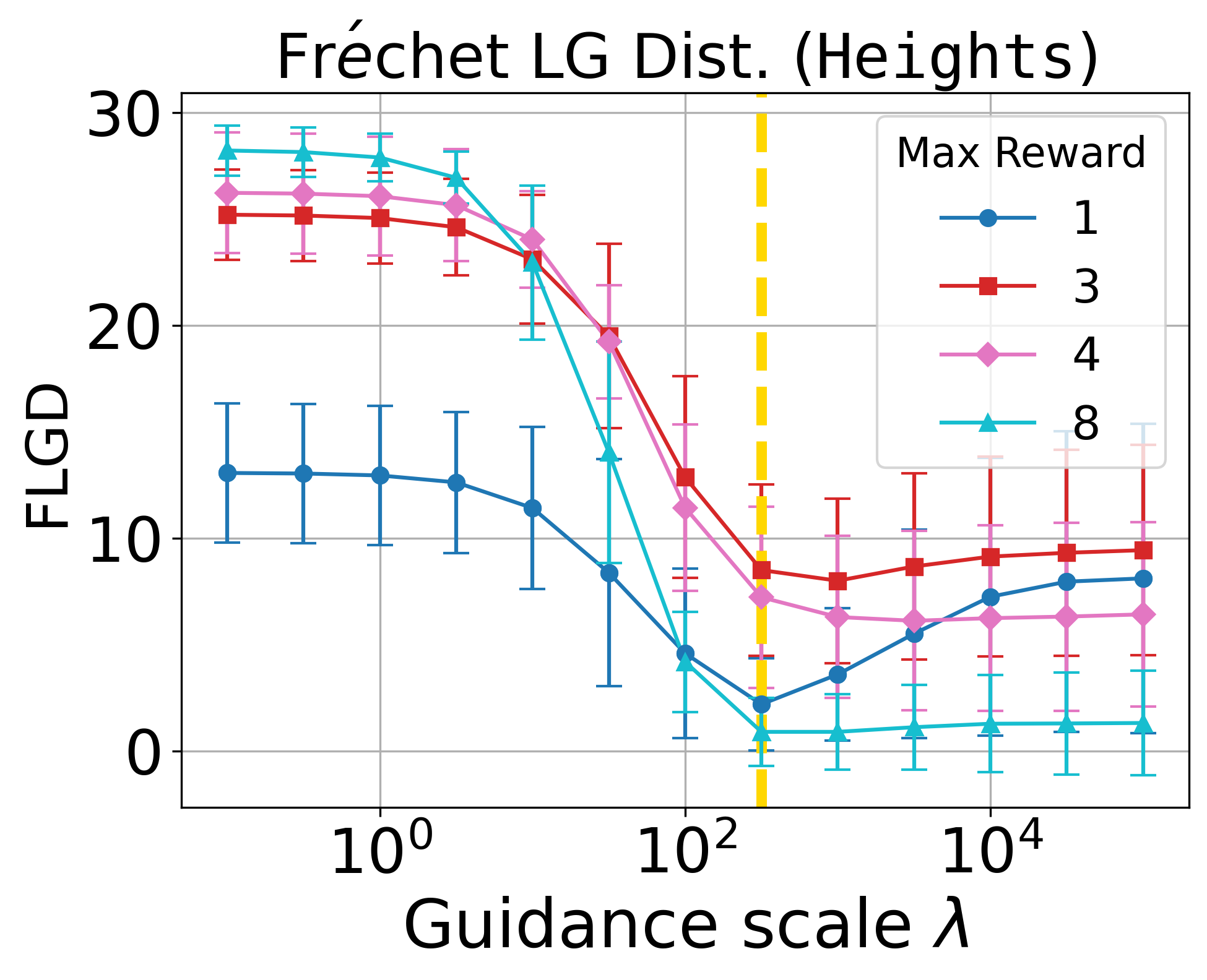}
        \label{fig:flgd_columbia}
    }
    \hfill
    \subfigure[FLGD $\boldsymbol{\downarrow}$ \texttt{Ridge}]{
        \includegraphics[width=.3\columnwidth]{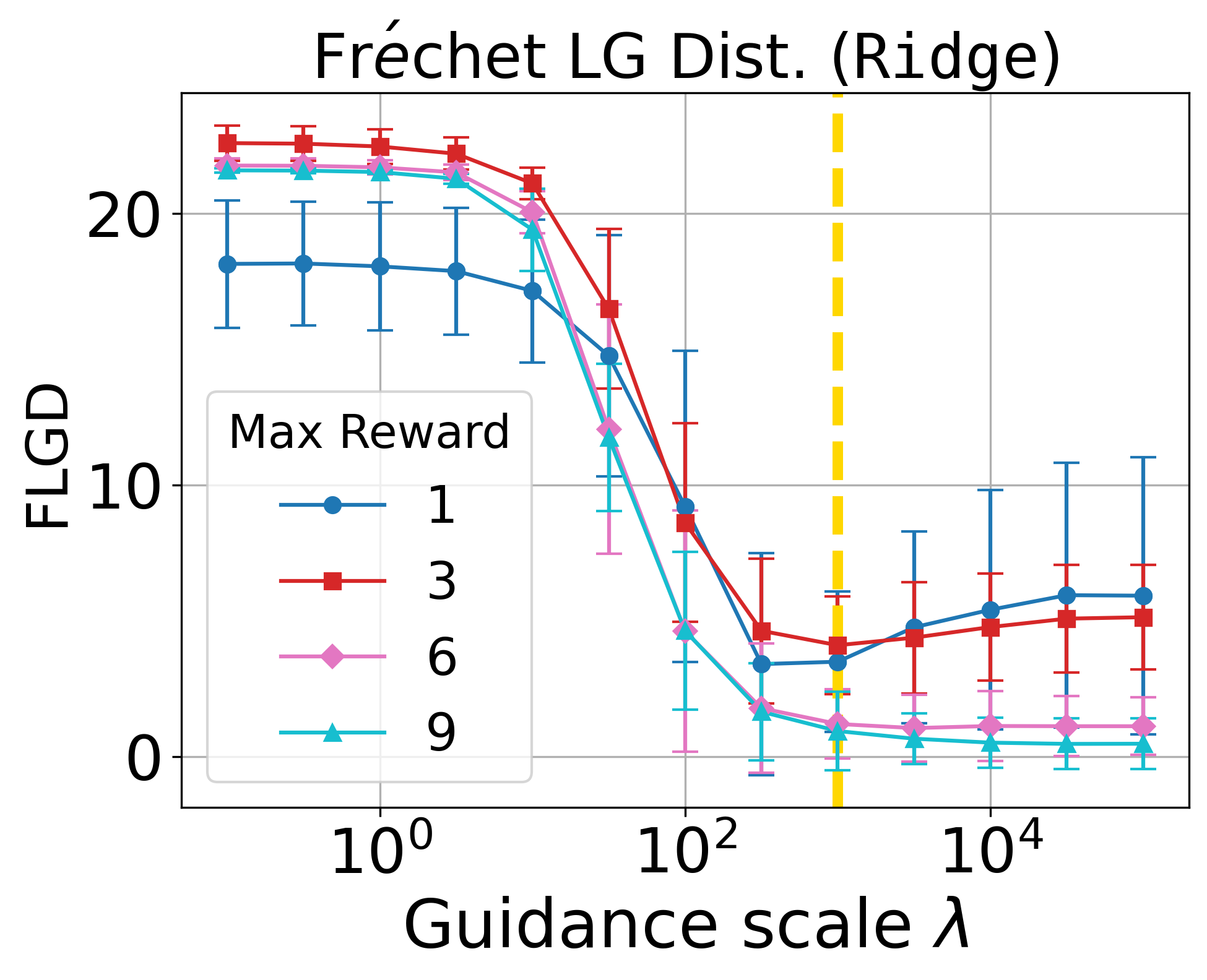}
        \label{fig:sfd_nus}
    }
    \hfill
    \caption{Additional experiments with guidance on \texttt{Toy}, \texttt{Heights} and \texttt{Ridge}. The first row contains the rewards obtained. The next 4 rows are based on FLGD and IS-L metrics. }
    \label{fig:scores_all}
    \end{figure}